\documentclass[a4paper,fleqn]{cas-sc}

\usepackage[authoryear]{natbib}

\def\tsc#1{\csdef{#1}{\textsc{\lowercase{#1}}\xspace}}
\tsc{WGM}
\tsc{QE}

\usepackage{hyperref}

\usepackage{times}  
\usepackage{helvet}  
\usepackage{courier}  
\usepackage{multirow}
\usepackage{amssymb}
\usepackage{multirow}
\usepackage{amssymb}
\usepackage{tikz}
\usepackage{algorithm}
\usepackage{algorithmic}
\usepackage{xcolor}
\usepackage{booktabs} 
\usepackage{tabularx}
\usepackage{amsmath}
\usepackage{mdframed}
\usepackage{array}
\usepackage{cleveref}
\usepackage{listings}
\usepackage{caption}
\usepackage{chngcntr}
\usepackage{subcaption}

\crefname{table}{Table}{Table}
\crefname{figure}{Fig.}{Fig.}
\crefname{equation}{Eq.}{Eq.}
\crefname{section}{Section}{Section}
\crefname{algorithm}{Algorithm.}{Algorithm.}
\crefname{section}{Section}{Section}

\newcommand{\perfup}[2]{#1\% \small$\uparrow$#2}
\newcommand{\perfdown}[2]{#1\% \small$\downarrow$#2}
\newcommand{\perfsame}[1]{#1\% \small- 0.00}

\usepackage{tcolorbox}
\tcbuselibrary{breakable}
\tcbset{colback=white,colframe=black!60,arc=2pt,boxrule=0.6pt}
\begin{document}
\let\WriteBookmarks\relax
\def\floatpagepagefraction{1}
\def\textpagefraction{.001}

\shorttitle{ORThought: Benchmarking and Automating Logistics Optimization Modeling}    

\shortauthors{Yang et al}  

\title [mode = title]{ORThought: Benchmarking and Automating Logistics Optimization Modeling via Structured LLM Reasoning}  



\author[1]{Beinuo Yang}[auid=000]
\ead{beinuo.24@intl.zju.edu.cn}
\credit{Conceptualization, Methodology, Software, Investigation, Formal analysis, Writing – original draft}

\author[2, 1]{Qishen Zhou}[auid=001]
\cormark[1]
\ead{qishenzhou@jlu.edu.cn}
\credit{Conceptualization, Methodology, Investigation, Formal analysis, Supervision, Writing – review \& editing}

\author[3]{Junyi Li}[auid=002]
\ead{junyi.li@smart.mit.edu}
\credit{Methodology, Investigation, Writing – review \& editing}

\author[4]{Chenxing Su}[auid=003]
\ead{scx@suchenxing.com}
\credit{Resources, Writing – review \& editing}

\author[5]{Panagiotis Angeloudis}[auid=004]
\ead{p.angeloudis@imperial.ac.uk}
\credit{Resources, Writing – review \& editing}

\author[1]{Simon Hu}[auid=005]
\cormark[1]
\ead{simonhu@zju.edu.cn}
\credit{Resources, Supervision, Writing – review \& editing}

\affiliation[a]{organization={ZJU-UIUC Institute, Zhejiang University},
            city={Haining},
            postcode={314400},
            country={China}}

\affiliation[b]{organization={School of Transportation, Jilin University},
            city={Changchun},
            postcode={130022},
            country={China}}

\affiliation[b]{organization={Singapore-MIT Alliance for Research and Technology},
            city={Singapore},
            postcode={138602}, 
            country={Singapore}}

\affiliation[c]{organization={Link.AI, Minimal Future Tech.},
            city={Shenzhen},
            postcode={518100},
            country={China}}

\affiliation[c]{organization={Department of Civil and Environmental Engineering, Imperial College London},
            city={London},
            postcode={SW7 2BU},
            country={UK}}

\cortext[1]{Corresponding authors: Qishen Zhou and Simon Hu}

\begin{abstract}
Optimization modeling stands as the engine of scientific decision-making in logistics and transportation, yet its adoption is hindered by a steep expertise threshold and the latency of manual workflows. Automating this process via Large Language Models (LLMs) offers a potential solution, but current approaches face critical bottlenecks: (i) a lack of high-quality, complex benchmarks; (ii) methodological inefficiencies in autonomous multi-agent frameworks, which often exhibit instability and redundant computation; and (iii) evaluations that lack diagnostic depth. In this work, we address these challenges from the following three aspects. First, we introduce LogiOR, a diverse logistics benchmark with rigorous annotations, and enrich existing datasets with the same annotation standard to support community utilization. Second, we propose ORThought, a structured dual-agent framework. By incorporating expert-level modeling principles via chain-of-thought reasoning, ORThought eliminates the redundancy of uncontrolled autonomous agents. Third, extensive empirical evaluations demonstrate that ORThought consistently outperforms state-of-the-art baselines by 9–17 percentage points, exhibiting distinct advantages in handling complex constraints while maintaining high token efficiency. Building on these results, we further conduct a multidimensional error analysis, which identifies key failure modes and success factors, providing actionable insights for future research. The dataset and code are available at \href{https://huggingface.co/datasets/LabMem012/LogiOR}{Hugging Face} and \href{https://github.com/ZJU-TSELab/ORThought}{GitHub}, respectively.
\end{abstract}

\begin{highlights}
\item Introducing a rigorous and comprehensively annotated benchmark for logistics optimization.

\item Proposing a structured dual-agent framework for Optimization modeling.

\item Achieving 9-17\% performance gains over SOTA on complex tasks.

\end{highlights}


\maketitle

\section{Introduction}

Operations Research (OR) serves as the cornerstone of scientific decision-making for modern logistics and transportation, driving efficiency across global supply chains~\citep{hillier2021IntroductionOR}. While decades of methodological advancement have yielded sophisticated algorithms capable of solving complex linear, integer, and stochastic problems efficiently, the critical step of transforming operational realities into mathematical programs, known as Optimization Modeling (OM), remains a persistent bottleneck dependent on human expertise.

This reliance on manual formulation imposes severe structural limitations. Firstly, it creates a prohibitive "expertise gap": mastering OM typically demands advanced academic training, thereby restricting adoption among practitioners~\citep{Gurobi2024report}. Secondly, it reduces responsiveness to dynamic conditions. In logistics operations, decision rules and constraints frequently change in response to volatile markets and real-time service requirements~\citep{richey2022responsiveness}. The latency of manual reformulation cannot keep pace with such evolving conditions, often rendering static models obsolete. Moreover, effective modeling necessitates bridging the communication gap between domain stakeholders and OR experts. This heavy communication overhead frequently results in misalignment and costly iterative cycles, making the manual workflow not only inaccessible and slow but also susceptible to coordination friction and iterative inefficiency. These challenges highlight the need for Automated Optimization Modeling (AOM) approaches that reduce the expertise burden, enable operational agility, and streamline the end-to-end modeling workflow.

\begin{figure}
    \centering
    \includegraphics[width=\textwidth]{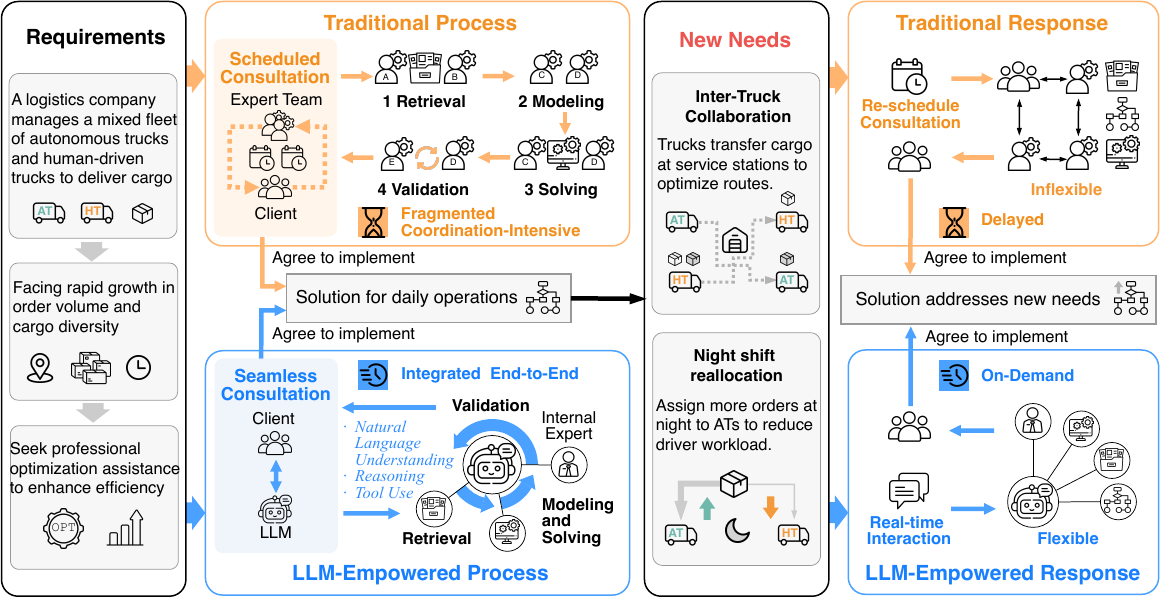}
    \caption{Comparison of workflows in logistics optimization modeling. The traditional expert-dependent workflow (top) suffers from high coordination latency and slow adaptation. In contrast, the LLM-empowered approach (bottom) integrates retrieval, modeling, and solving into a seamless natural language interface, enabling real-time responsiveness to dynamic operational needs.}
    \label{fig:example}
\end{figure}

Recent breakthroughs in LLMs, particularly in natural language understanding~\citep{brown2020GPT3}, mathematical reasoning~\citep{huang2023towards_reasoning_survey}, and formal code generation~\citep{llm_code}, are inherently aligned with the core demands of the OM process, making them promising enablers of AOM. Motivated by this potential, researchers have begun exploring LLMs for AOM, with early evidence emerging from the 2023 NL4Opt competition~\citep{ramamonjison2023NL4Opt}, where LLMs for the first time outperformed task-specific fine-tuned models in modeling linear programming problems. Building on this milestone, a rapidly growing body of literature has explored harnessing LLMs for AOM, achieving initial successes in canonical constrained optimization problems~\citep{mostajabdaveh2024optimization, ahmed2024lm4opt, deng2024CAFA} and subsequently extending to more complex scenarios involving lengthy textual descriptions~\citep{OptiMUS24ICML} and implicit constraints~\citep{xiao2023ChainofExperts}. More recently, this line of research has advanced to domain-specific challenges such as vehicle routing~\citep{jiang2025droc} and dynamic fleet management~\citep{jiang2025rideagent}. This growing body of work demonstrates the viability of LLM-based AOM across diverse problem settings. Building on these advances, we envision that, compared to the traditional expert-dependent workflow, an LLM-integrated AOM approach can significantly reduce formulation latency and communication overhead, with \cref{fig:example} providing a concrete example in the context of logistics operations.

Despite these initial successes, current research faces the following critical challenges: 
(i) Existing benchmarks often lack comprehensive annotations such as mathematical formulations and problem characteristics, hindering fine-grained analysis and reliable evaluation, and many focus on toy-size, structurally simplistic problems with limited complexity. Moreover, a dedicated benchmark for the logistics and transportation domain remains absent.
(ii) Recent trends favor complex, highly autonomous multi-agent frameworks. While novel, these systems often rely on uncontrolled inter-agent interactions, incurring substantial computational overhead and instability (e.g., cascading errors) without providing systematic evidence of superiority over simpler, structured approaches. 
(iii) While aggregate accuracy is commonly reported, existing studies often lack multidimensional assessments across varying problem sizes, complexities, and failure types. This absence of diagnostic evaluation obscures specific capability boundaries. To address these challenges, this paper makes the following contributions:

\begin{itemize}
    \item We introduce LogiOR, a rigorously annotated benchmark dedicated to logistics and transportation optimization, featuring greater problem complexity and diversity than prior datasets. Each instance is equipped with a complete mathematical formulation, executable Gurobi code, a solver-verified optimal objective value, and problem characteristics, enabling fine-grained error analysis and providing rich supervisory signals for training. We further apply the same annotation standard to three existing benchmarks, enriching their annotations to facilitate community utilization and verification.

    \item We propose ORThought, a resource-efficient structured framework. Unlike complex autonomous systems that rely on potentially emergent behaviors, ORThought employs a minimalist dual-agent architecture to embed expert-level modeling principles via structured chain-of-thought reasoning. This design achieves state-of-the-art modeling success rates while significantly reducing computational costs and communication overhead.

    \item We conduct a comprehensive evaluation across multiple benchmarks and foundation models. Results demonstrate that ORThought consistently outperforms existing approaches, including sophisticated autonomous frameworks and fine-tuned models, by 9–17 percentage points, with distinct advantages on complex, larger-size problems. Furthermore, through detailed ablation studies and error analyses, we identify key failure modes, providing actionable insights for future research into LLM-based optimization modeling.

\end{itemize} 

The remainder of this paper is organized as follows. Section 2 reviews related work. Section 3 introduces the proposed ORThought framework. Section 4 presents the LogiOR benchmark. Section 5 reports experimental evaluations and analysis. Section 6 discusses current limitations and future directions, followed by the conclusion in Section 7.

\section{Literature Review}

Recent advances in LLMs have introduced a suite of capabilities that are particularly well-suited to the demands of AOM. A foundational capability is in-context learning \citep{brown2020GPT3, dong2024incontext-learning-survey}, which allows LLMs to perform formulation without parameter updates, thereby significantly lowering the barrier to deployment. For the intricate work of translating natural language into precise mathematics, Chain-of-Thought (CoT) reasoning~\citep{wei2022chain-of-thought, chu2024cot_survey} constitutes a pivotal advancement. By decomposing problems into intermediate steps, CoT significantly enhances performance on logical tasks and provides a transparent window into the model's problem-solving process. Building upon this, structured reasoning frameworks such as Tree of Thoughts~\citep{yao2023tree-of-thoughts} and Graph of Thoughts~\citep{besta2024graph-of-thoughts} enable more sophisticated exploration of solution spaces through branching and backtracking, while Self-Consistency~\citep{wang2023self-consistency} improves reliability by aggregating multiple reasoning paths to reach robust conclusions. Self-reflection mechanisms, exemplified by Reflexion \citep{shinn2023reflexion} and Self-Refine \citep{madaan2023self-refine}, further enable dynamic error correction. These reasoning paradigms are productively integrated into LLM agent frameworks, which combine these reasoning capabilities with tool use \citep{qin2024toolllm}, transforming LLMs into proactive, multi-step problem-solvers, a paradigm well-aligned with the multi-stage nature of OM. In parallel, specialized training through techniques like supervised fine-tuning~\citep{ouyang2022instructGPT} offers a pathway to directly imbue LLMs with enhanced capabilities in specific domains such as mathematical reasoning and code generation. Building on these capabilities, a growing body of work has explored their application to AOM. We organize this review along three dimensions: existing benchmarks and evaluations, training-based methods that adapt model parameters, and inference-based methods that leverage in-context learning, structured reasoning, and agent frameworks. 

\subsection{Benchmarks and evaluation}
High-quality benchmarks are fundamental to method development and evaluation. \cite{ramamonjison2023NL4Opt} introduced the first optimization problem benchmark with natural language descriptions, NL4Opt, comprising linear programming problems from six domains, including sales and investment. \cite{xiao2023ChainofExperts} noted that real-world optimization problem descriptions contain implicit constraints and domain-specific terminology, and accordingly proposed the ComplexOR benchmark. \cite{OptiMUS24ICML} considered that real-world problem descriptions are typically longer, proposed NLP4LP, a long-text description benchmark containing both linear programming and mixed-integer linear programming problems. \cite{huang2025ORLM} directly collected real optimization problems from industrial practice to construct IndustryOR, a benchmark with higher complexity. \cite{yang2025optibench} proposed the OptiBench dataset, which incorporates tabular information and enriches problem types.

Existing benchmarks have three major limitations. First, existing benchmarks contain substantial annotation errors and incomplete problem descriptions~\citep{jiang2025LLMOPT}. Second, problem complexity is limited, with most benchmarks focusing on toy-size linear programming. Third, many benchmarks lack detailed annotations essential for analysis and training. These benchmark limitations lead to insufficient evaluations. Most studies rely solely on overall success rates, which mask performance variations across problem types, sizes, and categories. Moreover, computational costs are rarely evaluated, with scarcely any works reporting token consumption despite their practical importance. Additionally, few works examine the modeling process, making it difficult to identify specific failure points.

\subsection{Training-based methods}
Training-based methods seek to enhance LLMs for AOM by directly adapting their parameters via techniques such as supervised fine-tuning and reinforcement learning. The majority of existing literature focuses on improving the performance of smaller, more efficient models. For instance, \cite{ahmed2024lm4opt} proposed a progressive fine-tuning framework for this purpose, while \cite{huang2025ORLM} constructed a specialized dataset through a semi-automated workflow and applied supervised fine-tuning to produce ORLM, demonstrating that a finely-tuned small model can achieve performance competitive with larger foundation models. Other research has explored more structured training approaches. \cite{jiang2025LLMOPT} defined optimization problems using a five-element framework and employed multi-instruction fine-tuning to improve accuracy. Besides supervised learning, \cite{chen2025SIRL} introduced the SIRL framework, which uses external solvers to provide verification signals for reinforcement learning. Similarly, \cite{zhou2025autoformulatingDP} adopted a two-stage pipeline combining fine-tuning and RL for dynamic programming problems.

Despite the significant performance improvements, these methods face limitations in cost and transferability. Specifically, they typically require substantial computational resources for high-quality data curation and model training~\citep{liang2025LargeScaleLEAN}. Furthermore, the specialized capabilities they acquired are often tightly coupled with the specific model architectures, making them difficult to transfer across different foundation models and necessitating frequent retraining as foundation models advance. 

\subsection{Inference-based methods}
Inference-based methods elicit OM capabilities of LLMs without requiring parameter updates, primarily by leveraging in-context learning and agent frameworks. These methods can be categorized along a spectrum of increasing architectural complexity, each presenting distinct trade-offs between flexibility, controllability, and computational cost.

At one end of the spectrum are single-agent approaches, which enhance a single LLM's performance through carefully designed prompts or external algorithms, without inter-agent coordination. The CAFA \citep{deng2024CAFA}, for example, employs one compact prompt guiding the LLMs to transform problem descriptions into code, achieving high success rates on linear programming problems. Separately, \cite{astorga2025autoformulationMCTS} integrated Monte Carlo Tree Search an external algorithm to explore and optimize the model's reasoning paths hierarchically. While straightforward to implement, these approaches rely on a single model's capabilities without leveraging collaboration among specialized agents.

Moving beyond single-agent paradigms, structured multi-agent frameworks employ predefined pipelines where specialized agents sequentially execute subtasks. The ORMind framework \citep{wang2025ORMind}, inspired by cognitive theory, is a representative example. Subsequent works like \cite{liang2025LargeScaleLEAN} and \cite{thind2025OptimAI} have augmented this basic structure with retrieval-augmented generation and dedicated planner agents, respectively. The principal strength of these approaches lies in their high transparency and controllability due to the fixed workflow. However, this very rigidity can limit the system's adaptability when confronting novel or highly dynamic problem scenarios that require flexible strategy adjustments.

Pursuing greater flexibility, autonomous multi-agent frameworks implement dynamic coordination mechanisms, allowing agents to interact and determine execution paths with less predefinition. Frameworks such as Chain-of-Experts \citep{xiao2023ChainofExperts} and OptiMUS \citep{OptiMUS24ICML} exemplify this direction, employing conductor or manager agents to orchestrate specialized roles. The key advantage of these approaches is their ability to construct strategies for diverse and complex problems dynamically. This flexibility, however, is accompanied by substantial computational overhead from extensive inter-agent communication and an increased risk of cascading errors.

While these methods demonstrate increasing capability, they also introduce growing architectural complexity, raising concerns about computational overhead and system instability. Moreover, systematic comparisons with simpler baselines remain scarce, leaving the marginal benefit of such complexity largely underexplored.

\section{Methodology}

In this section, we present ORThought, a structured dual-agent framework that decomposes automated optimization modeling into two core capabilities: abstract mathematical formulation and concrete code implementation. This division is grounded in their distinct cognitive demands, with the former requiring logical reasoning about problem structure and mathematical relationships, while the latter necessitates precise debugging of implementation details and solver interactions.

\begin{figure}
    \centering
    \includegraphics[width=\textwidth]{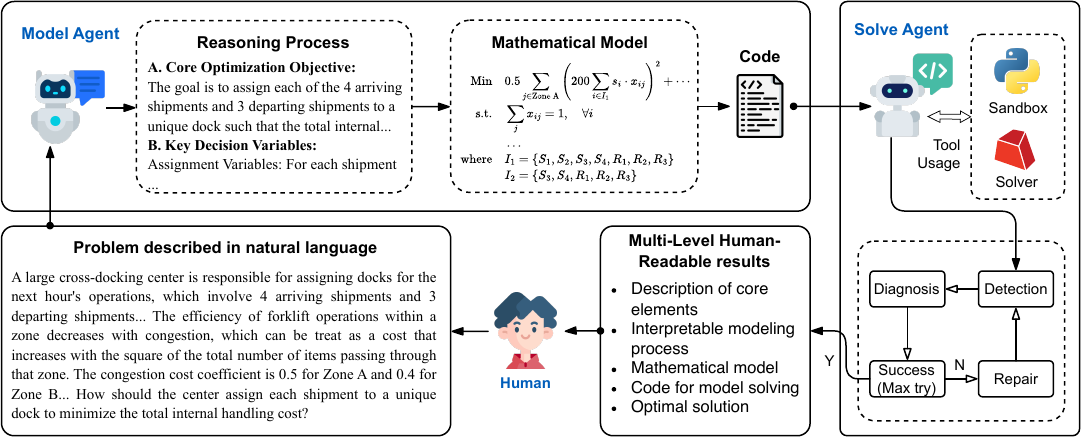}
    \caption{The framework of ORThought.}
    \label{fig:framework}
\end{figure}

To effectively address these dual requirements, ORThought employs a dual-agent architecture as illustrated in ~\cref{fig:framework}. The Model Agent, leveraging chain-of-thought reasoning grounded in expert-level modeling principles, comprehends real-world problems and translates them into precise mathematical models with corresponding solution code. The Solve Agent, equipped with a Python execution environment and advanced solvers such as Gurobi, proceeds to execute code and iteratively refine the solutions through an intelligent Detection-Diagnosis-Repair workflow. This separation of concerns is motivated by evidence that task switching, such as alternating between abstract mathematical reasoning and concrete code debugging, can degrade LLM performance \citep{gupta-etal-2024-task-switch}. By dedicating each agent to a focused cognitive task, ORThought mitigates this interference while maintaining clear reasoning contexts. This modular design further enables clear error attribution and facilitates component-wise optimization. 
Both agents are designed using prompts, which have been proven effective in guiding LLMs to perform complex reasoning tasks and enhancing their domain-specific problem-solving capabilities ~\citep{prompt_survey_liu2023PretrainPromptPredict}. Detailed prompts are provided in the Appendix. The following subsections detail their design and functionality.

\subsection{Model agent}

The Model Agent $\mathcal{A}_M$ transforms a natural language problem description into optimization code through a three-stage pipeline: problem understanding, mathematical modeling, and code generation. Let $Q$ denote the natural language problem description and $P_M$ the Model Agent's prompt. The Model Agent takes $Q$ as input and produces three structured outputs in sequence within a single inference call:

$$(\mathcal{U}, \mathcal{M}, \mathcal{K}^0) = \mathcal{A}_M(Q; P_M)$$
where $\mathcal{U}$ denotes the structured problem understanding, $\mathcal{M}$ the formal mathematical model, and $\mathcal{K}^0$ the executable Gurobi Python code. This pipeline leverages the natural cognitive flow of OM, where each stage produces verifiable intermediate outputs that guide subsequent ones, ensuring a systematic transformation while maintaining transparency and interpretability.

\paragraph{Problem understanding} 
In this first stage, $\mathcal{A}_M$ builds a structured understanding $\mathcal{U}$ of the problem $Q$. Following the OR modeling methodology established in \citep{book_williams2013modelbuilding_in_mathematical_programming}, the expert knowledge embedded in $P_M$ at this stage prescribes a structured decomposition with a specific identification order: objectives first, then decision variables, then constraints. This order reflects the principle that a clear understanding of the optimization goal should precede the definition of the decision space. Concretely, the agent distinguishes maximization from minimization objectives, classifies decision variables into continuous, integer, and binary types, and organizes constraints by their logical role (e.g., resource limitations, operational requirements). This disciplined approach ensures a comprehensive and unambiguous problem representation, avoiding the pitfalls of free-form interpretation and providing a solid foundation for mathematical formalization.

\paragraph{Mathematical modeling}

In this stage, $\mathcal{A}_M$ progressively constructs the formal mathematical model $\mathcal{M}$. The expert knowledge embedded in $P_M$ prescribes a formalization sequence that differs from the identification order above: decision variable domains are defined first, followed by term-by-term objective function construction with derivation, and finally constraint formalization with explicit logical justification. This ordering reflects the requirement that precise mathematical construction requires the decision space to be fully established before the objective function and constraints can be expressed rigorously. Throughout this process, the agent maintains clarity, interpretability, and notational consistency, and the final integration of all components using standardized optimization notation yields a complete, well-formed mathematical model $\mathcal{M}$ that accurately captures $Q$ and is immediately ready for codification. 

\paragraph{Code generation}
In the final stage, $\mathcal{A}_M$ generates executable Python code $\mathcal{K}^0$ conditioned on $Q$, $\mathcal{U}$, and $\mathcal{M}$, using the Gurobi solver interface \citep{gurobi}. We selected Gurobi for its robust performance and comprehensive documentation. The framework architecture, however, remains compatible with alternative solvers like CPLEX \citep{manual1987ibm} and open-source options. To enhance modularity and reusability, $\mathcal{K}$ is structured as a parameterized function that encapsulates the entire model logic. It adheres to a standardized return protocol: the optimal objective value for feasible solutions, and \texttt{None} for infeasible or unbounded cases. This explicit return scheme guarantees consistent reporting across diverse problem scenarios and enables seamless integration with automated evaluation pipelines.

Through these three stages, $\mathcal{A}_M$ systematically transforms $Q$ into $(\mathcal{U}, \mathcal{M}, \mathcal{K})$ within a single inference call. The integration of operations research expertise across $\mathcal{U}$ and $\mathcal{M}$ ensures that the generated models accurately capture 
problem requirements, with the interpretable intermediate representations providing clear checkpoints for error attribution.

\subsection{Solve agent}
The Solve Agent $\mathcal{A}_S$ functions as the execution and refinement engine within ORThought, employing an iterative three-phase workflow---Detection, Diagnosis, and Repair---to ensure robust solution quality. Let $P_S$ denote $\mathcal{A}_S$'s prompt. At each iteration $t$, $\mathcal{A}_S$ takes the current code $\mathcal{K}^t$, the mathematical model $\mathcal{M}$, and the problem description $Q$ as input, executes $\mathcal{K}^t$ to obtain execution feedback $\mathcal{E}^t = \text{Execute}(\mathcal{K}^t)$, and terminates if execution succeeds, or produces corrected code $\mathcal{K}^{t+1}$ otherwise:
$$\mathcal{K}^{t+1} = \mathcal{A}_S(Q, \mathcal{M}, \mathcal{K}^t, \mathcal{E}^t; P_S)$$
Unlike $\mathcal{A}_M$'s single-pass pipeline, $\mathcal{A}_S$ operates cyclically, repeatedly running the loop until either obtaining a valid solution or reaching predefined termination conditions.
It is important to note that the repair mechanism is triggered solely by execution failures, where $\mathcal{E}^t$ contains runtime exceptions or solver-reported errors. When the generated code executes successfully but the underlying mathematical model contains errors, $\mathcal{A}_S$ does not attempt to revise the formulation. Detecting and correcting such silent formulation errors would require formulation-level verification, involving consistency checks between $\mathcal{M}$ and $Q$, semantic validation of constraint logic, and reliable criteria for distinguishing correct from incorrect formulations, all of which remain open challenges in LLM-based AOM and are beyond the scope of the present work.
 
The design of $\mathcal{A}_S$ is motivated by two OR-specific considerations: (1) the architectural separation of debugging from formulation reasoning to avoid task-switching degradation \citep{gupta-etal-2024-task-switch}, and (2) the explicit conditioning of each repair cycle on $\mathcal{M}$ as a fidelity anchor, ensuring repairs do not drift from the intended optimization semantics.

\paragraph{Detection}
In this phase, $\mathcal{A}_S$ executes the generated Gurobi code $\mathcal{K}^t$ within a sandbox Python environment, capturing runtime information as execution feedback $\mathcal{E}^t = \text{Execute}(\mathcal{K}^t)$, where $\mathcal{E}^t$ is either a successful solver output or a set of error messages and stack traces. The detection mechanism primarily monitors for runtime exceptions (such as syntax errors, undefined variables, and type mismatches). $\mathcal{E}^t$ is then passed to the diagnosis phase for further analysis.

\paragraph{Diagnosis}
During this phase, $\mathcal{A}_S$ analyzes $\mathcal{E}^t$ when it indicates execution failure, with access to the complete context $(Q, \mathcal{M}, \mathcal{K}^t, \mathcal{E}^t)$. The agent examines 
the error logs and stack traces alongside $Q$, $\mathcal{M}$, and $\mathcal{K}^t$ to pinpoint the source of failure, which may stem from coding errors (e.g., incorrect API usage or variable naming issues), model formulation mistakes (e.g., constraint definition errors), or solver-reported conditions (e.g., numerical issues). The agent produces a brief diagnostic explanation of the identified issue, which directly guides the subsequent repair strategy.

\paragraph{Repair}
Upon error detection and diagnosis, the repair phase generates corrected code $\mathcal{K}^{t+1}$ that addresses the identified issues. Corrections may address variable definitions, constraint syntax, solver parameters, or errors in translating mathematical expressions to code (e.g., reversed inequality directions or oversimplified implementation of complex constraint logic). Crucially, the repair process maintains fidelity to the original mathematical formulation $\mathcal{M}$, ensuring fixes resolve implementation defects without altering the optimization model's structure or logic. The corrected code $\mathcal{K}^{t+1}$ is then passed back to the Detection phase, continuing the cycle until a termination condition is met.

The workflow iterates under clear termination conditions: (1) successful execution with the solver reporting an optimal solution, (2) successful execution with the solver indicating infeasibility or unboundedness, or (3) reaching the maximum iteration limit $T_{\max}$. This iterative refinement mechanism allows the framework to recover from initial implementation errors, boosting solution success rates beyond single-pass code generation.

\section{The LogiOR Benchmark}
\label{sec:logior_benchmark}
Existing benchmarks for AOM were not originally designed for the logistics and transportation domain, and tend to focus on simpler problem types with limited complexity and annotation depth. To bridge this gap, we construct LogiOR, a dedicated benchmark for logistics and transportation optimization, designed around the following three objectives: (1) to provide a dedicated benchmark for the logistics and transportation domain; (2) to encompass a diverse range of problem types and complexity levels, with particular emphasis on medium-scale instances, mixed-integer linear programming, and non-linear programming problems that are underrepresented in existing benchmarks; and (3) to offer comprehensive annotations that support fine-grained analysis and facilitate community utilization and verification.

\subsection{Dataset construction}

The construction of LogiOR proceeds in three phases, conducted over two months under the guidance of three operations research experts. First, problems are drawn from classical OR solver test datasets \citep{beasley1990or-library}, textbook examples \citep{book_williams2013modelbuilding_in_mathematical_programming, book_winston2004operations}, research papers, and real-world logistics applications, with natural language descriptions composed or rewritten to ensure clarity and to reduce overlap with problems that may have been encountered during LLM pretraining. Second, each problem is independently annotated by one of the three experts, yielding a complete mathematical formulation, executable Gurobi code, a solver-verified optimal objective value, and problem characteristics including type and size metrics. Third, the two experts not responsible for each instance independently review its annotation to ensure correctness before the instance is admitted to the dataset. An example problem from LogiOR is shown in Appendix~\ref{sec:logior}.

\subsection{Dataset analysis}

\begin{figure}
    \centering
    \includegraphics{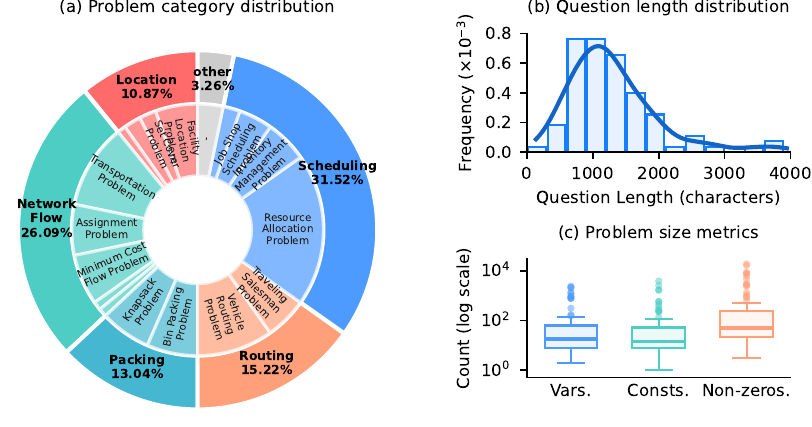}
    \caption{Detailed statistics of the LogiOR benchmark: (a) distribution of problem categories, (b) distribution of question text length, and (c) problem size metrics (variables, constraints, and non-zeros).}
    \label{fig:logior}
\end{figure}

As shown in \cref{fig:logior} and \cref{tab:data}, LogiOR exhibits diversity across problem categories, question lengths, and problem-size metrics. It covers six problem categories spanning a wide range of logistics scenarios. Mathematically, it includes Linear Programming (LP), Integer Linear Programming (ILP), Mixed-Integer Linear Programming (MILP), and Nonlinear Programming (NLP) problems. Problem sizes in terms of variables, constraints, and non-zeros span multiple orders of magnitude, reflecting varying degrees of computational complexity. Compared to existing benchmarks (see \cref{sec:eval_datasets} for their descriptions), LogiOR stands out with a higher proportion of challenging MILP and NLP problems, as well as a larger share of medium-scale instances. 

\begin{table}
\centering
    \begin{tabularx}{\textwidth}{l *{4}{>{\centering\arraybackslash}X}}
    \toprule
        & \textbf{LogiOR} & \textbf{ComplexOR} &\textbf{NLP4LP} & \textbf{IndustryOR} \\ \midrule
    LP     & 22     & 10        & 61     & 18         \\
    ILP     & 43     & 7         & 203    & 47         \\
    MILP    & 11     & 1         & 0      & 14         \\
    NLP     & 16      & 0         & 0      & 4          \\ \midrule
    Toy  & 14     & 13        & 264    & 33         \\
    Small & 39     & 5         & 0      & 36         \\
    Medium  & 39     & 0         & 0      & 14         \\ 
    \bottomrule
    \end{tabularx}
\caption{Characteristics of the LogiOR benchmark and existing datasets. The problem size is classified as Toy ($<$ 5 variables, 10 constraints, 20 non-zero coefficients), Small ($<$ 25 variables, 40 constraints, 80 non-zero coefficients), or Medium (otherwise).}
\label{tab:data}
\end{table}

\section{Experiments and Analysis}

In this section, extensive experiments are conducted to answer the following questions:

\begin{itemize}
\item \textbf{RQ1:} How does ORThought's AOM performance compare to existing methods?
\item \textbf{RQ2:} How does ORThought's performance vary across problems with different characteristics?
\item \textbf{RQ3:} What are the main types of errors made by ORThought?
\item \textbf{RQ4:} How do different components of ORThought contribute to its overall performance?
\item \textbf{RQ5:} How do different hyperparameters (LLM choice, model size, and temperature) affect ORThought's performance?
\item \textbf{RQ6:} How does ORThought perform across reasoning and non-reasoning models, and can it enable lightweight models to approach reasoning model performance?
\end{itemize}

\subsection{Datasets}
\label{sec:eval_datasets}

We evaluate ORThought on four benchmarks that collectively provide comprehensive coverage of optimization problem types and difficulty levels. These include our newly proposed LogiOR benchmark and three enhanced well-known existing datasets:

\begin{itemize} 

\item \textbf{ComplexOR.} The ComplexOR dataset \citep{xiao2023ChainofExperts}, originally developed with three OR experts, spans domains of supply chain, industry scheduling, and logistics. The original annotations include mathematical formulations, executable solver code, and optimal objective value. We utilized all 18 problems from its GitHub repository, corrected annotation errors in 6 instances, and enhanced the dataset by adding problem type and size annotations following the LogiOR annotation standard.

\item \textbf{NLP4LP.} The NLP4LP dataset \citep{ramamonjison2023NL4Opt} comprises optimization problems across retail, energy, and other industrial domains. The original annotations include executable solver code and optimal objective value. From its HuggingFace repository, we selected all 269 problems, removed 5 problems with insufficient information for modeling, corrected annotation errors in 44 problems, and enriched the dataset with mathematical formulations, problem type, and size annotations following the LogiOR standard.

\item \textbf{IndustryOR.} The IndustryOR dataset \citep{huang2025ORLM}, the first industrial dataset specifically designed for optimization modeling, contains 100 real-world OR problems from eight sectors, including education, transportation, and finance. The original annotations only include the optimal objective value. We enhanced the dataset by removing 17 problems with insufficient information, correcting annotation errors in 23 problems, and adding mathematical formulations, executable solver code, problem type, and size annotations following the LogiOR standard.

\end{itemize}

\subsection{Experiment setup}

Unless otherwise specified, GPT-4.1-nano~\citep{GPT4} with temperature 0 is used as the backbone. Solve agent's maximum iteration limit is set to $T_{\max}=3$. Each experiment is repeated three times to account for potential randomness. We use average success rate as the primary evaluation metric, where a trial is considered successful if the LLM-generated solution achieves the ground-truth optimal objective value verified by OR experts. Additionally, we measure computational efficiency by tracking the average token consumption of each method.

We compare our approach against four categories of baselines: a supervised fine-tuned model ORLM-LLaMA-3-8b (ORLM) \citep{huang2025ORLM}, autonomous multi-agent methods including Chain-of-Experts (CoE) \citep{xiao2023ChainofExperts} and OptiMUS \citep{OptiMUS24ICML}; reasoning methods including Chain-of-Thought (CoT) ~\citep{wei2022chain-of-thought}, Self-Consistency (SC) ~\citep{wang2023self-consistency}, and Reflexion \citep{shinn2023reflexion}; and a vanilla baseline that directly generates the mathematical model with a simple prompt. For ORLM, we deploy the model open-sourced by the authors on Hugging Face Inference Endpoint. For CoE and OptiMUS, we strictly follow their official implementations, with minor adaptations to CoE's code to ensure a fair comparison. This controlled setup ensures that performance differences reflect framework design rather than model capability differences.

\subsection{Overall performance (RQ1)}

\subsubsection{Success rate}

As shown in \cref{tab:overall_results}, ORThought consistently outperforms existing methods across all datasets, with relative improvements of 13-28 percentage points over the Standard baseline. Notably, it outperforms even the most competitive reasoning methods (Reflexion and CoT) by 9-17 percentage points. The performance varies significantly across datasets: while achieving a high success rate of 89.02\% on NLP4LP, the performance decreases on more complex datasets (57.83\% on IndustryOR and 46.01\% on LogiOR). These results demonstrate the effectiveness of ORThought while also revealing the persistent challenges in handling complex optimization problems.

\begin{table}
    \centering
    \begin{tabularx}{\textwidth}{XXXXX}
        \toprule
        \textbf{Method} & \textbf{NLP4LP (264)} & \textbf{IndustryOR (83)} & \textbf{LogiOR (92)} & \textbf{ComplexOR (18)} \\
        \midrule
        Standard & 72.73\% & 42.17\% & 33.34\% & 50.00\% \\
        \midrule
        ORLM & \perfdown{67.8}{4.93}\% & \perfdown{32.13}{10.04}\% & \perfdown{15.58}{17.76}\% & \perfup{77.78}\textbf{{27.7}8}\% \\
        OptiMUS & \perfdown{69.42}{3.31} & / & / & / \\
        CoE & \perfup{75.00}{2.27} & \perfdown{40.96}{1.21} & \perfup{34.78}{1.44} & \perfup{55.56}{5.56} \\
        CoT & \perfup{75.00}{2.27} & \perfup{43.37}{1.20} & \perfup{37.32}{\underline{3.98}} & \perfup{61.11}{\underline{11.11}} \\
        SC & \perfdown{71.21}{1.52} & \perfup{48.19}{\underline{6.02}} & \perfdown{32.25}{1.09} & \perfsame{50.00} \\
        Reflexion & \perfup{77.65}{\underline{4.92}} & \perfup{48.19}{\underline{6.02}} & \perfup{36.59}{3.25} & \perfup{61.11}{\underline{11.11}} \\
        ORThought (ours) & \perfup{89.02}{\textbf{16.29}} & \perfup{57.83}{\textbf{15.66}} & \perfup{46.01}{\textbf{12.67}} & \perfup{77.78}{\textbf{27.78}} \\
        \bottomrule
    \end{tabularx}
    \caption{Comparison of success rates across different methods on various datasets. Numbers in parentheses indicate sample size. Bold: best performance; Underline: second best. For each method, the first number shows the success rate, while arrows ($\uparrow$/$\downarrow$) followed by numbers indicate percentage points increase/decrease compared to the Standard baseline.}
    \label{tab:overall_results}
\end{table}

A detailed examination of different methodologies reveals distinct patterns. The supervised fine-tuned model ORLM exhibits limited performance, with noticeable improvement only on ComplexOR, while underperforming the Standard baseline on the other three datasets. Specifically, on the more complex IndustryOR and LogiOR datasets, ORLM's performance decreases by 10.04 and 17.76 percentage points, respectively. This performance gap can be attributed to several factors. First, while ORLM was originally reported to rival GPT-4, it is based on a fine-tuned 8B-parameter model, whereas other methods in our experiment utilize the more recent and powerful GPT-4.1-nano, representing a substantial disparity in model capacity. This also highlights potential challenges of supervised fine-tuning approaches, where training outcomes may have limited transferability and struggle to keep pace with rapidly evolving foundation models. Additionally, ORLM's training data was synthesized through a semi-automatic pipeline, which may contain noise and errors that could impact model performance.

Among multi-agent approaches, OptiMUS\footnote{OptiMUS requires an intermediate step of converting problems in natural language into JSON format, which is treated as an integral part of its overall pipeline. As this conversion consistently produces JSON parsing errors when using GPT-4.1-nano as the backbone, we only evaluated OptiMUS on the NLP4LP dataset using the authors' provided JSON outputs to bypass this problematic transformation step.} shows degraded performance compared to the Standard baseline, with a drop of 3.31 percentage points on NLP4LP. This deterioration can be attributed to its workflow-based design, where errors cascade through sequential agent interactions. While CoE mitigates this cascading error problem through a shared information pool, its unconstrained agent collaboration often leads to illogical operation sequences, such as code generation before mathematical model formulation, limiting its performance improvement.

Reasoning methods generally demonstrate better performance than multi-agent approaches, with Reflexion and CoT emerging as the most competitive baseline methods across different datasets. While we do not view these results as definitive evidence of the superiority of reasoning methods over multi-agent approaches, they do highlight that reasoning methods are easier to design and implement, and still achieve competitive performance in optimization modeling tasks.  ORThought builds upon these insights by incorporating expert-level planning into the reasoning process, achieving 9-17 percentage points improvements over these strong baselines while maintaining implementation simplicity.

\begin{figure}
    \centering
    \includegraphics[]{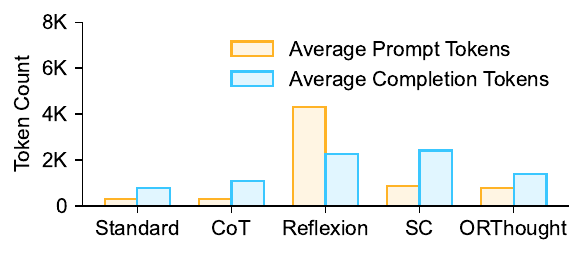}
    \caption{Token consumption comparison across methods.}
    \label{fig:overall_tokens}
\end{figure}
\subsubsection{Token cost} 
To evaluate computational efficiency, \cref{fig:overall_tokens} analyzes the average token consumption per problem across all datasets. The average prompt token represents the average number of input tokens for task instructions and examples, while the average completion token indicates the average number of output tokens generated by the model. ORLM is not included in the figure as it uses a different tokenizer (LLaMA3-8b). For reference, ORLM's token consumption measured with its own tokenizer is 229.12 prompt tokens and 1022.02 completion tokens on average. 

Autonomous multi-agent methods like CoE and OptiMUS are excluded from the figure due to their inherently higher token usage granted by free exploration privileges - for instance, CoE's average prompt token reaches around 50K. Among reasoning methods,  ORThought demonstrates low average prompt token usage, slightly higher than CoT, and significantly lower than Reflexion, the method achieves the second-best accuracy. Its average completion token is slightly higher than CoT but significantly lower than SC and Reflexion. These results demonstrate that  ORThought achieves superior performance while maintaining excellent token efficiency.

\subsection{Performance analysis by problem characteristics (RQ2)}

\begin{table}
    \centering

    \begin{tabularx}{\textwidth}{l *{4}{>{\centering\arraybackslash}X}}
        \toprule
        \textbf{Method} & \textbf{LP (111)} & \textbf{ILP (300)} & \textbf{MILP (26)} & \textbf{NLP (20)} \\
        \midrule
        Standard & 62.16\% & 66.00\% & 19.23\% & 13.35\% \\
        CoE & 56.76\% & \underline{67.67\%} & 19.23\% & 15.00\% \\
        CoT & \underline{66.67\%} & 64.67\% & 23.08\% & \underline{26.65\%} \\
        Reflexion & \textbf{68.47}\% & 67.00\% & \underline{28.19\%} & \underline{26.65}\% \\
        ORThought & \underline{66.67\%} & \textbf{82.33\%} & \textbf{30.77\%} & \textbf{51.65\%} \\
        \midrule
        Improve & $\downarrow$ 1.80 & $\uparrow$ 14.66 & $\uparrow$ 2.58 & $\uparrow$ 25.00 \\
        \bottomrule
    \end{tabularx}
    \caption{Comparison of success rates across different optimization problem types. Improve: percentage points gained by ORThought compared to the most competitive baseline.}
    \label{tab:by_type}
\end{table}

\begin{table}
    \centering
    \begin{tabularx}{\textwidth}{l *{3}{>{\centering\arraybackslash}X}}
        \toprule
        \textbf{Method} & \textbf{Toy (324)} & \textbf{Small (80)} & \textbf{Medium (53)} \\
        \midrule
        Standard & 68.93\% & 36.66\% & 26.42\% \\
        CoE & 69.44\% & 40.00\% & \underline{32.08\%} \\
        CoT & 70.68\% & 41.66\% & \underline{32.08\%} \\
        Reflexion & \underline{74.07\%} & \underline{43.34\%} & 28.30\% \\
        ORThought & \textbf{85.39\%} & \textbf{49.59\%} & \textbf{43.40\%} \\
        \midrule
        Imporve & $\uparrow$ 11.32 & $\uparrow$ 6.25  & $\uparrow$ 11.32 \\
        \bottomrule
    \end{tabularx}
    \caption{Comparison of success rates across different optimization problem sizes.}
    \label{tab:by_size}

\end{table}

To better understand ORThought's advantages, we analyze its performance across different problem types, sizes, and categories. As shown in Table \ref{tab:by_type}, ORThought significantly outperforms baselines on three out of four problem types, achieving the highest success rate on ILP, MILP, and NLP problems. For LP problems, ORThought matches the second-best performance while falling 1.80 percentage points behind Reflexion. The performance gains are particularly pronounced for ILP and NLP problems, where ORThought improves upon the second-best methods by 14.66 and 25.00 percentage points, respectively. This suggests that ORThought's structured reasoning approach is especially effective for more challenging problem types.

Table \ref{tab:by_size} further reveals ORThought's performance across different problem sizes. ORThought consistently achieves the highest success rates, with notable improvements of 11.32, 6.25, and 11.32 percentage points over the second-best methods for toy, small, and medium problems, respectively. Examining performance across problem sizes reveals a consistent pattern: success rates decline sharply as problem complexity increases, with all methods showing significant performance degradation from toy to medium problems. This universal trend underscores that scaling to larger optimization problems remains a fundamental challenge in the field, pointing to a critical direction for future research.

\begin{table}
    \centering
    \begin{tabularx}{\textwidth}{l *{6}{>{\centering\arraybackslash}X}}
        \toprule
        \textbf{Method} & \textbf{Location} & \textbf{Net. Flow} & \textbf{Packing} & \textbf{Routing} & \textbf{Scheduling} & \textbf{Other} \\
        & \textbf{(10)} & \textbf{(24)} & \textbf{(12)} & \textbf{(14)} & \textbf{(29)} & \textbf{(3)} \\
        \midrule
        Standard & 43.33\% & 33.33\% & 50.00\% & 9.52\% & 37.93\% & 0.00\% \\
        CoE & 46.67\% & \underline{37.50\%} & 50.00\% & \underline{14.29\%} & 35.63\% & 0.00\% \\
        CoT & \textbf{56.67\%} & \textbf{43.06\%} & 41.67\% & \underline{14.29\%} & 37.93\% & \underline{11.11}\% \\
        Reflexion & 43.33\% & 34.72\% & \underline{66.67\%} & 0.00\% & \textbf{44.83\%} & 0.00\% \\
        ORThought & \underline{50.00\%} & \underline{37.50\%} & \textbf{88.89\%} & \textbf{21.43\%} & \textbf{44.83\%} & \textbf{55.56\%} \\
        \midrule
        Improve & $\downarrow$ 6.67 & $\downarrow$ 5.56 & $\uparrow$ 22.22 & $\uparrow$ 7.14 & $\uparrow$ 0.00 & $\uparrow$ 44.44 \\
        \bottomrule
    \end{tabularx}
    \caption{Comparison of success rates across different categories on the LogiOR dataset. Net. Flow: Network Flow.}
    \label{tab:by_application}
\end{table}

To further understand performance variations, Table \ref{tab:by_application} presents results across different categories in the LogiOR dataset. Despite limited sample sizes in some categories, several patterns emerge: ORThought achieves substantial gains in four of six categories, with its most pronounced advantages emerging in complex combinatorial domains like Packing, where it improves upon the second-best method by 22.22 percentage points. The modest decreases in highly structured categories such as Location and Network Flow (6.67 and 5.56 percentage points below the best methods, respectively) suggest that for problems with well-established canonical formulation templates, the marginal benefit of guided reasoning may be reduced. Collectively, these results indicate that ORThought's primary strength lies in tackling problems that require sophisticated modeling decisions beyond the application of standardized formulations.

Collectively, these results demonstrate that ORThought's structured reasoning approach provides particular advantages for complex problem types (ILP, MILP, NLP) and larger problem instances, while maintaining competitive performance on simpler problems where established formulation patterns exist.

\subsection{Error analysis (RQ3)}

\begin{figure}
    \centering
    \includegraphics{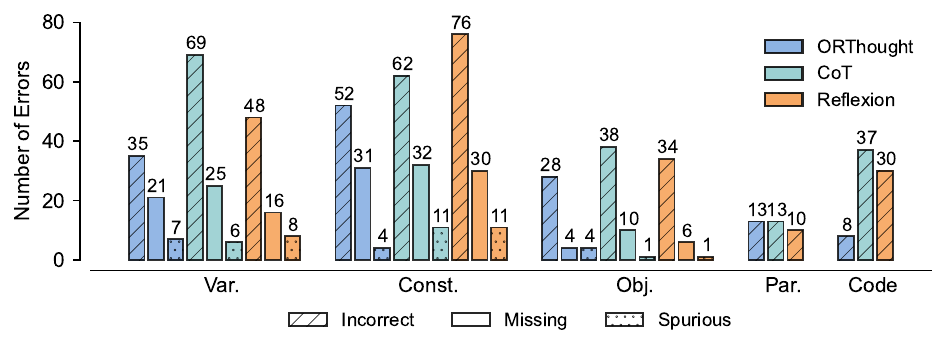}
    \caption{Analysis of modeling failures in ORThought, Chain-of-Thought (CoT) and Reflexion across different components. Here, Var., Obj., Const., Par., and Code represent variables, objective functions, constraints, parameters in the optimization model, and Gurobi Python code, respectively.}
    \label{fig:detailed_error_comparison}
\end{figure}
We analyze error patterns across ORThought, CoT, and Reflexion, the two strongest baselines in terms of overall success rate. Our fine-grained error analysis (\cref{fig:detailed_error_comparison}) reveals distinct failure patterns along two dimensions: error types and optimization elements. Notably, errors are not mutually exclusive, a single failed problem may contain multiple errors across different elements or exhibit multiple error types simultaneously, and our analysis counts all identified errors. We classify modeling errors into three types based on their nature. The incorrect type refers to cases where the LLM correctly identifies the need for a certain element but formulates it improperly. The missing type captures necessary elements that are entirely omitted, while the spurious type refers to falsely introduced elements that should not exist. For the objective function specifically, missing and spurious errors refer to absent or extraneous terms within the objective expression rather than the objective function itself being omitted or duplicated.

Across all three methods, constraints are consistently the most error-prone component, followed by variables and objective functions. Within each component, incorrect errors predominate over missing and spurious errors, indicating that LLMs more often attempt to model elements but do so improperly than omit them entirely or introduce baseless ones. Missing errors show little variation across methods, suggesting that exhaustively capturing all necessary problem information remains a shared challenge. These shared patterns suggest that the fundamental challenges of translating complex logical relationships into precise mathematical expressions remain a key bottleneck in LLM-based AOM.

Beyond these shared patterns, the comparison reveals several notable differences. The number of errors in variables and objective functions is inversely correlated with overall success rates across the three methods, with ORThought incurring the fewest and CoT the most, highlighting the critical role of accurately specifying these elements. ORThought's explicit structured decomposition of decision variables and objective functions during the problem understanding and mathematical modeling stages likely contributes to its advantage here. ORThought also incurs substantially fewer code errors than CoT and Reflexion, which is especially noteworthy given that Reflexion employs an execution feedback mechanism for error correction yet still exhibits considerably more code errors, suggesting that structured reasoning and the dedicated separation of modeling and solving concerns may be more effective than post-hoc recovery within a single agent.

\subsection{Ablation study (RQ4)}

\begin{figure}
    \centering
    \includegraphics{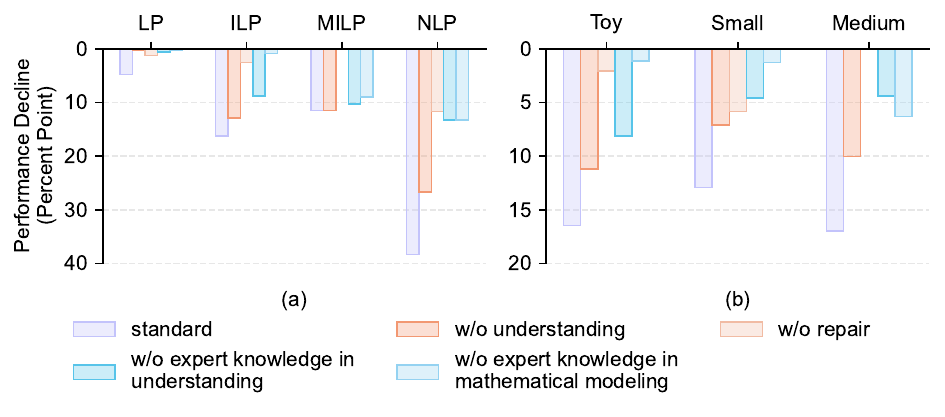}
    \caption{Impact of different components in ORThought across (a) problem types and (b) problem sizes.}
    \label{fig:ablation_study_by_type_and_size}
\end{figure}

To systematically evaluate the contribution of different components in ORThought, we conduct four ablation experiments: (1) removing the understanding module in Model Agent, (2) removing the repair functionality in Solve Agent, (3) removing expert knowledge in the understanding module in Model Agent, and (4) removing expert knowledge in the mathematical modeling module in Model Agent. These experiments target two dimensions: architectural modules (1–2) and expert knowledge integration at different reasoning stages (3–4). Results comparing each variant against the full framework are shown in \Cref{fig:ablation_study_by_type_and_size}. The standard baseline is also included as a reference. Detailed experimental settings for each ablation variant can be found in Appendix \ref{sec:appendix_c}.

First, we evaluate the impact of complete module removal. The removal of the understanding module in Model Agent led to significant performance degradation across all problem types except LP, with performance drops particularly pronounced in more complex problems, confirming its role as a cornerstone for successful AOM. The repair functionality shows a relatively modest contribution. However, it is worth noting that this repair mechanism only activates when the LLM-generated solution code fails to execute. Therefore, we conducted a dedicated evaluation of code execution errors and found that this functionality successfully corrected execution errors in 25 out of 31 problems, with 11 of these ultimately yielding the correct optimal solution, underscoring its indispensable role in system robustness.

We further examine the contribution by removing expert knowledge from the understanding and mathematical modeling modules, respectively. The results show that expert knowledge substantially boosts performance in the understanding phase, and while its impact on mathematical modeling is generally smaller, it demonstrates a comparable or even stronger impact in NLP problems and medium-size instances, revealing its crucial value in handling complex scenarios.

For LP problems, while ORThought consistently outperforms the standard baseline, the contribution of individual components is less distinctive. This suggests that ORThought's superior performance in LP problems stems from the synergistic effect of its components rather than any single module, warranting further investigation into the underlying mechanisms.

These results reveal a graded contribution pattern: the understanding module provides essential foundational analysis across all problem types, expert knowledge yields increasing benefits as problem complexity grows, and the repair mechanism ensures system robustness by recovering from execution failures. 

\begin{table}
    \centering
    \begin{tabularx}{\textwidth}{l *{4}{>{\centering\arraybackslash}X}}
        \toprule
        \textbf{Method} & \textbf{GPT-4.1-nano} & \textbf{DeepSeek-V3} & \textbf{Qwen3-32b} \\
        \midrule
        Standard & 58.35\% & 58.75\% & 64.55\% \\
        \midrule
        CoT & \perfup{62.12}{3.77} & \perfup{59.96}{1.21} & \perfup{67.83}{3.28} \\
        SC & \perfsame{58.35} & \perfdown{55.47}{3.28} & \perfup{68.93}{\underline{4.38}} \\
        Reflexion & \perfup{63.39}{\underline{5.04}} & \perfup{65.86}{\underline{7.11}} & \perfup{68.27}{3.72} \\
        ORThought & \perfup{74.25}\textbf{{15.90}} & \perfup{71.01}\textbf{{12.26}} & \perfup{73.38}{\textbf{8.83}} \\
        \bottomrule
    \end{tabularx}
    \caption{Performance across Different LLMs.}
    \label{tab:robustness}
\end{table}

\subsection{Hyperparameter analysis (RQ5)}
\subsubsection{LLM choice}
We evaluate ORThought across three representative LLMs: DeepSeek-V3 \citep{DeepSeekV3}, Qwen3-32B \citep{Qwen3}, and GPT4.1 Nano. As shown in Table \ref{tab:robustness}, the standard approach results demonstrate inherent differences in optimization modeling capabilities across base LLMs. However, ORThought effectively bridges these gaps, achieving consistently high performance across all models. Notably, our method maintains superior performance regardless of the underlying LLM, outperforming the second-best baseline by a substantial margin (4-10 percentage points).

\subsubsection{LLM size}

\begin{figure}
    \centering    
    \includegraphics{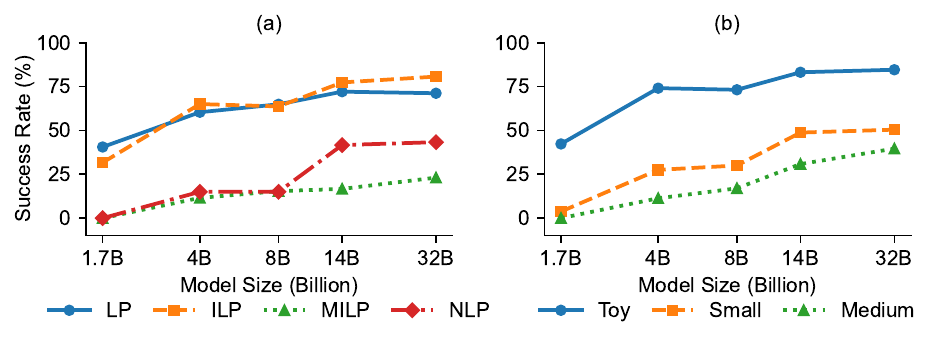}
    \caption{Performance of ORThought under Different LLM Model Sizes (x-axis in log scale).}
    \label{fig:model_scaling}
\end{figure}

We conduct experiments using the Qwen3 model series, spanning from 1.7B to 32B parameters, to investigate how ORThought's solving capabilities evolve across model sizes. As shown in \cref{fig:model_scaling}, our experiments reveal several key findings:

From the perspective of problem types, simpler problems (LP and ILP) show strong performance even with smaller models. Their success rates increase most rapidly in the early scaling stages (1.7B to 4B), followed by continued but more moderate improvements at larger sizes. In contrast, more complex problems (MILP and NLP) are unsolvable by small models and show slow initial improvement, but exhibit a notable performance leap at the 8-14B size.

Problem size analysis reveals similar scaling behaviors. Toy-size problems show rapid early improvements followed by gradual gains. Small-size problems demonstrate a distinct performance leap at 8-14B, while medium-size problems show consistent improvements with model size, indicating that larger models are particularly beneficial for handling increased problem complexity.

Two notable plateaus emerge in the scaling curves: one at 4-8B and another at 14-32B. These plateaus, combined with the observed performance leaps, suggest that certain reasoning capabilities may emerge at specific parameter thresholds rather than scaling smoothly with model size.

These results establish clear relationships between model size and AOM capabilities, revealing that performance improvements do not scale uniformly across problem types and sizes. This insight suggests that in practice, model selection should be guided by matching the problem's complexity to the model that has surpassed the requisite capability threshold, rather than unconditionally opting for the largest available size.

\begin{figure}
    \centering
    \includegraphics{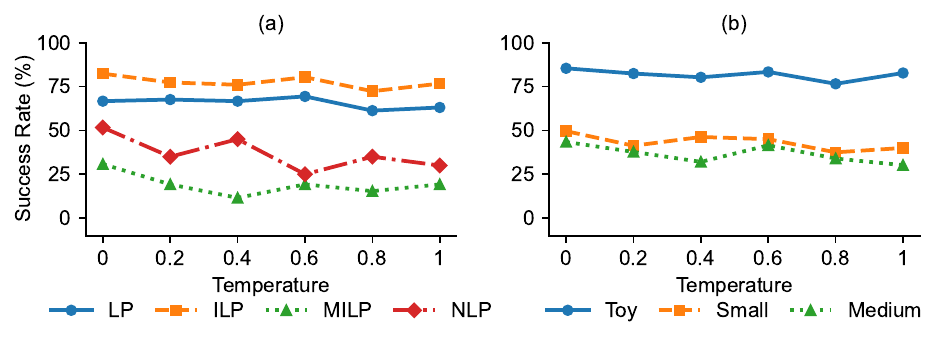}
    \caption{Impact of LLM temperature on ORThought performance.}
    \label{fig:temperature}
\end{figure}

\subsubsection{Temperature}

As shown in \cref{fig:temperature}, we examine the effect of LLM temperature on ORThought's performance across the range [0,1], analyzing different problem types and sizes. The results reveal varying sensitivity to LLM temperature changes across different categories. For problem types except MILP, despite some fluctuations in the curves, the overall trend shows declining performance with higher temperatures. This general downward trend is also observed across different problem sizes. MILP problems exhibit a distinct U-shaped pattern, with performance reaching its lowest point at temperature 0.4 before showing slight recovery. Despite these variations, temperature 0 yields the best performance for most problem categories, with LP problems showing relatively stable performance across temperatures from 0 to 0.6. These results suggest that deterministic LLM generation generally produces more reliable optimization modeling outcomes.

\subsection{Reasoning model (RQ6)} 

\begin{figure}
    \centering
    \includegraphics{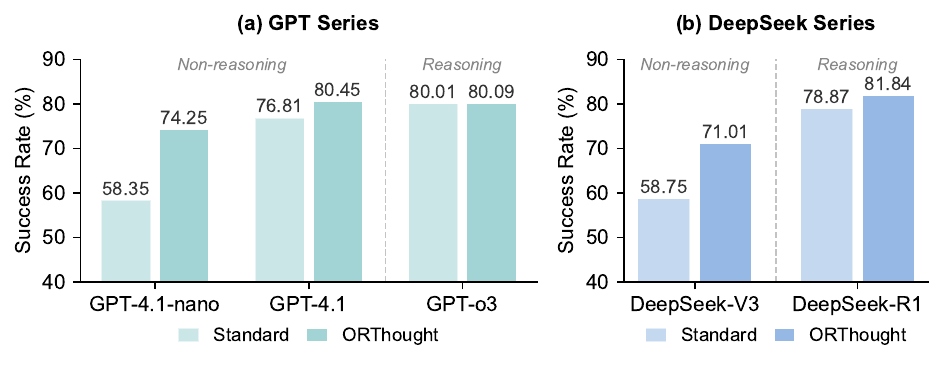}
    \caption{Performance comparison between Standard prompting and ORThought across non-reasoning and reasoning models.}
    \label{fig:reasoning_comparison}
\end{figure}

We evaluate ORThought and standard prompting on three non-reasoning models (GPT-4.1-nano, GPT-4.1, DeepSeek-V3) and two reasoning models (GPT-o3, DeepSeek-R1), to explore whether ORThought can elevate lightweight LLMs to a level comparable with reasoning models, and whether it remains beneficial when applied to models that already possess strong reasoning capabilities. 
Results are averaged across four benchmarks and reported in \cref{fig:reasoning_comparison}.

The results show a clear contrast between model types. For non-reasoning models, ORThought delivers substantial improvements across all scales, with gains of 15.90 pp on GPT-4.1-nano, 3.64 pp on GPT-4.1, and 12.26 pp on DeepSeek-V3. Notably, ORThought-augmented GPT-4.1-nano and DeepSeek-V3 achieve over 90\% of the performance of GPT-o3 and DeepSeek-R1 under standard prompting, at substantially lower inference cost. For reasoning models, ORThought provides only marginal gains (approximately 3 pp improvement on DeepSeek-R1). We attribute this to two factors. On one hand, reasoning models under standard prompting already achieve high absolute accuracy, leaving limited and highly challenging room for further improvement. On the other hand, we acknowledge this as a limitation of the current approach: ORThought's structured guidance was designed to compensate for the weaker problem-decomposition capabilities of non-reasoning models, and therefore be largely redundant with the built-in reasoning capabilities that reasoning models already possess through their internal inference-time computation. How to design prompting strategies that are genuinely complementary to the internal reasoning of stronger models remains an open question for future work.

\section{Discussion}
\label{sec:discussion}

This work introduces ORThought, a structured dual-agent framework for Automated Optimization Modeling (AOM), and demonstrates its effectiveness across diverse logistics operation tasks. While the results validate the promises of LLM-based AOM, our analysis also reveals several limitations that outline important directions for future research.

First, ORThought performs single-pass formulation generation and does not refine the mathematical model once it is produced. The key blind spot of this design is the silent failure scenario, where an incorrect but executable formulation yields a numerically plausible yet wrong objective value, which the evaluation pipeline can not detect. We identify two complementary directions for overcoming this limitation. The first is offline iterative formulation learning, where the system accumulates verified modeling insights through continual learning cycles, progressively improving formulation quality across problem instances (e.g., \citealp{kong2026alphaopt}). The key challenge in this direction is ensuring the quality and correctness of the accumulated modeling insights, while also determining whether and when to invoke them at test time. The second is online iterative formulation verification, where the system actively checks whether the generated formulation is consistent with the problem description through both external solver signals (e.g., infeasibility certificates, dual variables) and LLM-based self-verification. This direction shares the spirit of evolutionary coding agents such as AlphaEvolve \citep{novikov2025alphaevolve}, which combine LLM-driven proposal with automated evaluators in an iterative refinement loop. The key challenge here is defining clear correctness criteria for formulation.

Second, the evaluation methods in this field are not comprehensive. The current evaluation focuses primarily on solution correctness while ignoring computational tractability. However, in practice, the efficiency of the model is equally crucial. For instance, two formulations might yield identical optimal values yet differ drastically in solving time due to structural properties like constraint tightness or symmetry. Therefore, the future evaluation frameworks should measure efficiency through solver runtime, memory consumption, and how performance scales with problem size. Such comprehensive assessments will favor formulations that are not only mathematically valid but also computationally advantageous, better reflecting the dual emphasis on solution quality and solving speed in real-world optimization.

Third, there is a distinct trade-off between resource efficiency and model capability. Our empirical results indicate that smaller LLMs can achieve a competitive success rate on simpler problem instances, suggesting that uniformly deploying large models incurs unnecessary computational overhead. This observation highlights a clear opportunity for adaptive model selection, where problems are routed to LLMs of appropriate size based on their estimated problem difficulty. A key challenge in implementing such a system is designing reliable difficulty estimators. Possible indicators include characteristics such as the convexity of the problem, the density of constraints, and the linearity of the objective function. Equally critical is establishing a mapping between difficulty levels and LLMs of different sizes. Future research could explore learning-based routers trained on historical problem-performance data to realize this routing process, enabling cost-effective AOM that balances success rate against computational budget.

Finally, the pursuit of fully automated modeling warrants critical re-examination. While 100\% automation is a compelling technical goal, it may not be the optimal operational paradigm for high-stakes logistics decisions. Real-world problems often involve unstated business contexts, changing preferences, or "soft constraints" that are difficult to encode explicitly. In such cases, a fully automated black-box system may struggle to win user trust. A more pragmatic path forward is Human-AI Collaboration, where the LLM acts as a "copilot", drafting initial formulations and offering suggestions, while the human expert retains responsibility for verification and final decisions. This paradigm shifts the goal from replacing experts to augmenting their capabilities, combining the efficiency of AI with the nuanced judgment of humans.

\section{Conclusion}

Optimization modeling stands as the engine of scientific decision-making in logistics and transportation, yet its broader adoption has been hindered by a steep expertise threshold and the latency of manual workflows. This work bridges this gap by introducing an LLM-based framework to automate the OM process.

We address critical obstacles in this domain through three key contributions. First, we construct LogiOR, a dedicated logistics benchmark that enhances problem complexity and diversity compared to existing datasets. Second, we propose ORThought, a structured dual-agent framework that embeds expert-level modeling principles into the generation process. Third, extensive experiments across four datasets demonstrate that ORThought achieves substantial performance gains of 9–17 percentage points over strong baselines while maintaining high token efficiency. These advantages prove consistent across various problem types, sizes, and base LLMs, underscoring the robustness and generalizability of our approach. Furthermore, we critically analyze existing limitations and identify iterative formulation verification, adaptive model selection, and Human-AI collaboration as pivotal directions for future research.

\printcredits
\section*{Declaration of competing interest}
The authors declare that they have no known competing financial interests or personal relationships that could have appeared to influence the work reported in this paper.

\section*{Acknowledgements}
This study was supported in part by the National Natural Science Foundation of China (72350710798, 52131202), State Key Laboratory (SKL) of Biobased Transportation Fuel Technology, ZJU-YST joint research center for fundamental science, the Sustainable Smart Liveable Cities Research Centre, Zhejiang Province, Zhejiang University Global Partnership Fund, the Zhejiang University Sustainable Smart Livable Cities Alliance (SSLCA).

\bibliographystyle{cas-model2-names}

\bibliography{cas-refs}

@book{hillier2021IntroductionOR,
  title = {Introduction to {{Operations Research}}},
  author = {Hillier, Frederick S.},
  year = 2021,
  month = jun,
  urldate = {2025-09-09},
  langid = {english}
}

@techreport{Gurobi2024report,
  author    = {{Gurobi Optimization, LLC}},
  title     = {State of Mathematical Optimization 2024},
  institution = {Gurobi Optimization, LLC},
  year      = {2024},
  url       = {https://www.gurobi.com/resources/report-state-of-mathematical-optimization-2024/},
  note      = {Accessed: 2025-11-02}
}

@article{brown2020GPT3, 
title={Language Models are Few-Shot Learners}, 
volume={33}, 
url={https://proceedings.neurips.cc/paper_files/paper/2020/hash/1457c0d6bfcb4967418bfb8ac142f64a-Abstract.html}, 
journal={Advances in Neural Information Processing Systems}, 
author={Brown, Tom and Mann, Benjamin and Ryder, Nick and Subbiah, Melanie and Kaplan, Jared D and Dhariwal, Prafulla and Neelakantan, Arvind and Shyam, Pranav and Sastry, Girish and Askell, Amanda and Agarwal, Sandhini and Herbert-Voss, Ariel and Krueger, Gretchen and Henighan, Tom and Child, Rewon and Ramesh, Aditya and Ziegler, Daniel and Wu, Jeffrey and Winter, Clemens and Hesse, Chris}, 
year={2020}, 
pages={1877-1901} 
}

@inproceedings{dong2024incontext-learning-survey,
    title = "A Survey on In-context Learning",
    author = "Dong, Qingxiu  and
      Li, Lei  and
      Dai, Damai  and
      Zheng, Ce  and
      Ma, Jingyuan  and
      Li, Rui  and
      Xia, Heming  and
      Xu, Jingjing  and
      Wu, Zhiyong  and
      Chang, Baobao  and
      Sun, Xu  and
      Li, Lei  and
      Sui, Zhifang",
    editor = "Al-Onaizan, Yaser  and
      Bansal, Mohit  and
      Chen, Yun-Nung",
    booktitle = "Proceedings of the 2024 Conference on Empirical Methods in Natural Language Processing",
    month = nov,
    year = "2024",
    address = "Miami, Florida, USA",
    publisher = "Association for Computational Linguistics",
    url = "https://aclanthology.org/2024.emnlp-main.64/",
    doi = "10.18653/v1/2024.emnlp-main.64",
    pages = "1107--1128"
}

@article{llm_code,
author = {Jiang, Juyong and Wang, Fan and Shen, Jiasi and Kim, Sungju and Kim, Sunghun},
title = {A Survey on Large Language Models for Code Generation},
year = {2025},
publisher = {Association for Computing Machinery},
address = {New York, NY, USA},
issn = {1049-331X},
url = {https://doi.org/10.1145/3747588},
doi = {10.1145/3747588},
note = {Just Accepted},
journal = {ACM Trans. Softw. Eng. Methodol.},
month = jul
}

@inproceedings{huang2023towards_reasoning_survey,
    title = "Towards Reasoning in Large Language Models: A Survey",
    author = "Huang, Jie  and Chang, Kevin Chen-Chuan",
    booktitle = "Findings of the Association for Computational Linguistics: ACL 2023",
    month = jul,
    year = "2023",
    address = "Toronto, Canada",
    publisher = "Association for Computational Linguistics",
    url = "https://aclanthology.org/2023.findings-acl.67/",
    doi = "10.18653/v1/2023.findings-acl.67",
    pages = "1049--1065",
}

@article{prompt_survey_liu2023PretrainPromptPredict,
  title = {Pre-Train, {{Prompt}}, and {{Predict}}: {{A Systematic Survey}} of {{Prompting Methods}} in {{Natural Language Processing}}},
  shorttitle = {Pre-Train, {{Prompt}}, and {{Predict}}},
  author = {Liu, Pengfei and Yuan, Weizhe and Fu, Jinlan and Jiang, Zhengbao and Hayashi, Hiroaki and Neubig, Graham},
  year = {2023},
  month = jan,
  journal = {ACM Comput. Surv.},
  volume = {55},
  number = {9},
  pages = {195:1--195:35},
  issn = {0360-0300},
  doi = {10.1145/3560815},
  urldate = {2025-07-16}
}

@article{wei2022chain-of-thought,
  title={Chain-of-thought prompting elicits reasoning in large language models},
  author={Wei, Jason and Wang, Xuezhi and Schuurmans, Dale and Bosma, Maarten and Xia, Fei and Chi, Ed and Le, Quoc V and Zhou, Denny and others},
  journal={Advances in neural information processing systems},
  volume={35},
  pages={24824--24837},
  year={2022}
}

@article{yao2023tree-of-thoughts,
  title={Tree of thoughts: Deliberate problem solving with large language models},
  author={Yao, Shunyu and Yu, Dian and Zhao, Jeffrey and Shafran, Izhak and Griffiths, Tom and Cao, Yuan and Narasimhan, Karthik},
  journal={Advances in neural information processing systems},
  volume={36},
  pages={11809--11822},
  year={2023}
}

@inproceedings{besta2024graph-of-thoughts,
  title={Graph of thoughts: Solving elaborate problems with large language models},
  author={Besta, Maciej and Blach, Nils and Kubicek, Ales and Gerstenberger, Robert and Podstawski, Michal and Gianinazzi, Lukas and Gajda, Joanna and Lehmann, Tomasz and Niewiadomski, Hubert and Nyczyk, Piotr and others},
  booktitle={Proceedings of the AAAI conference on artificial intelligence},
  volume={38},
  pages={17682--17690},
  year={2024}
}

@inproceedings{wang2023self-consistency,
title={Self-Consistency Improves Chain of Thought Reasoning in Language Models},
author={Xuezhi Wang and Jason Wei and Dale Schuurmans and Quoc V Le and Ed H. Chi and Sharan Narang and Aakanksha Chowdhery and Denny Zhou},
booktitle={The Eleventh International Conference on Learning Representations },
year={2023},
url={https://openreview.net/forum?id=1PL1NIMMrw}
}

@article{madaan2023self-refine,
  title={Self-refine: Iterative refinement with self-feedback},
  author={Madaan, Aman and Tandon, Niket and Gupta, Prakhar and Hallinan, Skyler and Gao, Luyu and Wiegreffe, Sarah and Alon, Uri and Dziri, Nouha and Prabhumoye, Shrimai and Yang, Yiming and others},
  journal={Advances in Neural Information Processing Systems},
  volume={36},
  pages={46534--46594},
  year={2023}
}

@article{shinn2023reflexion,
  title={Reflexion: Language agents with verbal reinforcement learning},
  author={Shinn, Noah and Cassano, Federico and Gopinath, Ashwin and Narasimhan, Karthik and Yao, Shunyu},
  journal={Advances in Neural Information Processing Systems},
  volume={36},
  pages={8634--8652},
  year={2023}
}

@inproceedings{gupta-etal-2024-task-switch,
    title = "{LLM} Task Interference: An Initial Study on the Impact of Task-Switch in Conversational History",
    author = "Gupta, Akash  and
      Sheth, Ivaxi  and
      Raina, Vyas  and
      Gales, Mark  and
      Fritz, Mario",
    editor = "Al-Onaizan, Yaser  and
      Bansal, Mohit  and
      Chen, Yun-Nung",
    booktitle = "Proceedings of the 2024 Conference on Empirical Methods in Natural Language Processing",
    month = nov,
    year = "2024",
    address = "Miami, Florida, USA",
    publisher = "Association for Computational Linguistics",
    url = "https://aclanthology.org/2024.emnlp-main.811/",
    doi = "10.18653/v1/2024.emnlp-main.811",
    pages = "14633--14652"
}

@inproceedings{qin2024toolllm,
title={Tool{LLM}: Facilitating Large Language Models to Master 16000+ Real-world {API}s},
author={Yujia Qin and Shihao Liang and Yining Ye and Kunlun Zhu and Lan Yan and Yaxi Lu and Yankai Lin and Xin Cong and Xiangru Tang and Bill Qian and Sihan Zhao and Lauren Hong and Runchu Tian and Ruobing Xie and Jie Zhou and Mark Gerstein and dahai li and Zhiyuan Liu and Maosong Sun},
booktitle={The Twelfth International Conference on Learning Representations},
year={2024},
url={https://openreview.net/forum?id=dHng2O0Jjr}
}

@book{book_williams2013modelbuilding_in_mathematical_programming,
  address = {Hoboken, N.J.},
  author = {Williams, H. P.},
  isbn = {9781118443330 1118443330},
  publisher = {Wiley},
  refid = {1153021717},
  title = {Model building in mathematical programming},
  url = {https://ebookcentral.proquest.com/lib/uvic/detail.action?docID=1120846},
  year = 2013
}

@book{book_winston2004operations,
  title={Operations research: applications and algorithm},
  author={Winston, Wayne L},
  year={2004},
  publisher={Thomson Learning, Inc.}
}

@inproceedings{xiao2023ChainofExperts,
  title = {Chain-of-{{Experts}}: {{When LLMs Meet Complex Operations Research Problems}}},
  shorttitle = {Chain-of-{{Experts}}},
  booktitle = {The {{Twelfth International Conference}} on {{Learning Representations}}},
  author = {Xiao, Ziyang and Zhang, Dongxiang and Wu, Yangjun and Xu, Lilin and Wang, Yuan Jessica and Han, Xiongwei and Fu, Xiaojin and Zhong, Tao and Zeng, Jia and Song, Mingli and Chen, Gang},
  year={2024},
  url={https://api.semanticscholar.org/CorpusID:271745931}
}

@inproceedings{ramamonjison2023NL4Opt,
  title = {{{NL4Opt Competition}}: {{Formulating Optimization Problems Based}} on {{Their Natural Language Descriptions}}},
  shorttitle = {{{NL4Opt Competition}}},
  booktitle = {Proceedings of the {{NeurIPS}} 2022 {{Competitions Track}}},
  author = {Ramamonjison, Rindranirina and Yu, Timothy and Li, Raymond and Li, Haley and Carenini, Giuseppe and Ghaddar, Bissan and He, Shiqi and Mostajabdaveh, Mahdi and {Banitalebi-Dehkordi}, Amin and Zhou, Zirui and Zhang, Yong},
  year = {2023},
  month = aug,
  pages = {189--203},
  publisher = {PMLR},
  issn = {2640-3498}
}

@inproceedings{deng2024CAFA,
  title = {{{CAFA}}: {{Coding}} as {{Auto-Formulation Can Boost Large Language Models}} in {{Solving Linear Programming Problem}}},
  shorttitle = {{{CAFA}}},
  booktitle = {The 4th {{Workshop}} on {{Mathematical Reasoning}} and {{AI}} at {{NeurIPS}}'24},
  author = {Deng, Haoxuan and Zheng, Bohao and Jiang, Yirui and Tran, Trung Hieu},
  year = {2024},
  month = oct
}

@article{liang2025LargeScaleLEAN,
  title = {{{LLM}} for {{Large-Scale Optimization Model Auto-Formulation}}: {{A Lightweight Few-Shot Learning Approach}}},
  author = {Liang, Kuo and Lu, Yuhang and Mao, Jianming and Sun, Shuyi and Yang, Chunwei and Zeng, Congcong and Jin, Xiao and Qin, Hanzhang and Zhu, Ruihao and Teo, Chung-Piaw},
  langid = {english},
  year = {2025},
}

@inproceedings{yang2025optibench,
title={OptiBench Meets ReSocratic: Measure and Improve {LLM}s for Optimization Modeling},
author={Zhicheng Yang and Yiwei Wang and Yinya Huang and Zhijiang Guo and Wei Shi and Xiongwei Han and Liang Feng and Linqi Song and Xiaodan Liang and Jing Tang},
booktitle={The Thirteenth International Conference on Learning Representations},
year={2025},
url={https://openreview.net/forum?id=fsDZwS49uY}
}

@InProceedings{OptiMUS24ICML,
  title = 	 {{O}pti{MUS}: Scalable Optimization Modeling with ({MI}){LP} Solvers and Large Language Models},
  author =       {Ahmaditeshnizi, Ali and Gao, Wenzhi and Udell, Madeleine},
  booktitle = 	 {Proceedings of the 41st International Conference on Machine Learning},
  pages = 	 {577--596},
  year = 	 {2024},
  editor = 	 {Salakhutdinov, Ruslan and Kolter, Zico and Heller, Katherine and Weller, Adrian and Oliver, Nuria and Scarlett, Jonathan and Berkenkamp, Felix},
  volume = 	 {235},
  series = 	 {Proceedings of Machine Learning Research},
  month = 	 {21--27 Jul},
  publisher =    {PMLR},
  url = 	 {https://proceedings.mlr.press/v235/ahmaditeshnizi24a.html}
}

@misc{wang2025ORMind,
  title = "{ORM}ind: A Cognitive-Inspired End-to-End Reasoning Framework for Operations Research",
    author = "Wang, Zhiyuan  and
      Chen, Bokui  and
      Huang, Yinya  and
      Cao, Qingxing  and
      He, Ming  and
      Fan, Jianping  and
      Liang, Xiaodan",
    editor = "Rehm, Georg  and
      Li, Yunyao",
    booktitle = "Proceedings of the 63rd Annual Meeting of the Association for Computational Linguistics (Volume 6: Industry Track)",
    month = jul,
    year = "2025",
    address = "Vienna, Austria",
    publisher = "Association for Computational Linguistics",
    url = "https://aclanthology.org/2025.acl-industry.10/",
    doi = "10.18653/v1/2025.acl-industry.10",
    pages = "104--131",
    ISBN = "979-8-89176-288-6",
}

@misc{thind2025OptimAI,
      title={OptimAI: Optimization from Natural Language Using LLM-Powered AI Agents}, 
      author={Raghav Thind and Youran Sun and Ling Liang and Haizhao Yang},
      year={2025},
      eprint={2504.16918},
      archivePrefix={arXiv},
      primaryClass={cs.CL},
      url={https://arxiv.org/abs/2504.16918}, 
}

@misc{astorga2025autoformulationMCTS,
      title={Autoformulation of Mathematical Optimization Models Using {LLM}s},
      author={Nicol{\'a}s Astorga and Tennison Liu and Yuanzhang Xiao and Mihaela van der Schaar},
      booktitle={Forty-second International Conference on Machine Learning},
      year={2025},
      url={https://openreview.net/forum?id=33YrT1j0O0}
}

@article{ahmed2024lm4opt,
  title={LM4OPT: Unveiling the potential of Large Language Models in formulating mathematical optimization problems},
  author={Tasnim Ahmed and Salimur Choudhury},
  journal={INFOR: Information Systems and Operational Research},
  year={2024},
  volume={62},
  pages={559-572},
  url={https://api.semanticscholar.org/CorpusID:268247666}
}

@article{huang2025ORLM,
  title = {{{ORLM}}: {{A Customizable Framework}} in {{Training Large Models}} for {{Automated Optimization Modeling}}},
  shorttitle = {{{ORLM}}},
  author = {Huang, Chenyu and Tang, Zhengyang and Hu, Shixi and Jiang, Ruoqing and Zheng, Xin and Ge, Dongdong and Wang, Benyou and Wang, Zizhuo},
  year = {2025},
  month = may,
  journal = {Operations Research},
  publisher = {INFORMS},
  issn = {0030-364X},
  doi = {10.1287/opre.2024.1233}
}

@inproceedings{jiang2025LLMOPT,
  title     = {LLMOPT: Learning to Define and Solve General Optimization Problems from Scratch},
  author    = {Caigao Jiang and Xiang Shu and Hong Qian and Xingyu Lu and Jun Zhou and Aimin Zhou and Yang Yu},
  booktitle = {Proceedings of the Thirteenth International Conference on Learning Representations (ICLR)},
  year      = {2025},
  address   = {Singapore, Singapore},
  url       = {https://openreview.net/pdf?id=9OMvtboTJg}
}

@article{chen2025SIRL,
  title={Solver-Informed RL: Grounding Large Language Models for Authentic Optimization Modeling},
  author={Chen, Yitian and Xia, Jingfan and Shao, Siyu and Ge, Dongdong and Ye, Yinyu},
  journal={arXiv preprint arXiv:2505.11792},
  year={2025}
}

@misc{zhou2025autoformulatingDP,
      title={Auto-Formulating Dynamic Programming Problems with Large Language Models}, 
      author={Chenyu Zhou and Jingyuan Yang and Linwei Xin and Yitian Chen and Ziyan He and Dongdong Ge},
      year={2025},
      eprint={2507.11737},
      archivePrefix={arXiv},
      primaryClass={cs.AI},
      url={https://arxiv.org/abs/2507.11737}, 
}

@misc{Qwen3,
  title = {{{Qwen3 Technical Report}}},
  author = {{Qwen Team}},
  year = {2025},
  month = may,
  number = {arXiv:2505.09388},
  eprint = {2505.09388},
  primaryclass = {cs},
  publisher = {arXiv},
  doi = {10.48550/arXiv.2505.09388},
  archiveprefix = {arXiv}
}

@misc{DeepSeekV3,
  title = {{{DeepSeek-V3 Technical Report}}},
  author = {{DeepSeek-AI}},
  year = {2025},
  month = feb,
  number = {arXiv:2412.19437},
  eprint = {2412.19437},
  primaryclass = {cs},
  publisher = {arXiv},
  doi = {10.48550/arXiv.2412.19437},
  archiveprefix = {arXiv}
}

@misc{GPT4,
  title = {{{GPT-4 Technical Report}}},
  author = {OpenAI},
  year = {2024},
  month = mar,
  number = {arXiv:2303.08774},
  eprint = {2303.08774},
  primaryclass = {cs},
  publisher = {arXiv},
  doi = {10.48550/arXiv.2303.08774},
  archiveprefix = {arXiv}
}

@article{beasley1990or-library,
  title={OR-Library: distributing test problems by electronic mail},
  author={Beasley, John E},
  journal={Journal of the operational research society},
  volume={41},
  number={11},
  pages={1069--1072},
  year={1990},
  publisher={Taylor \& Francis}
}

@misc{gurobi,
  author = {{Gurobi Optimization, LLC}},
  title = {{Gurobi Optimizer Reference Manual}},
  year = 2024,
  url = "https://www.gurobi.com"
}

@article{manual1987ibm,
  title={Ibm ilog cplex optimization studio},
  author={Manual, CPLEX User’s},
  journal={Version},
  volume={12},
  number={1987-2018},
  pages={1},
  year={1987}
}

@inproceedings{chu2024cot_survey,
  title={Navigate through enigmatic labyrinth a survey of chain of thought reasoning: Advances, frontiers and future},
  author={Chu, Zheng and Chen, Jingchang and Chen, Qianglong and Yu, Weijiang and He, Tao and Wang, Haotian and Peng, Weihua and Liu, Ming and Qin, Bing and Liu, Ting},
  booktitle={Proceedings of the 62nd Annual Meeting of the Association for Computational Linguistics (Volume 1: Long Papers)},
  pages={1173--1203},
  year={2024}
}

@article{richey2022responsiveness,
  title={A responsiveness view of logistics and supply chain management},
  author={Richey, Robert Glenn and Roath, Anthony S and Adams, Frank G and Wieland, Andreas},
  journal={Journal of Business Logistics},
  volume={43},
  number={1},
  pages={62--91},
  year={2022},
  publisher={Wiley Online Library}
}

@article{mostajabdaveh2024optimization,
  title={Optimization modeling and verification from problem specifications using a multi-agent multi-stage LLM framework},
  author={Mostajabdaveh, Mahdi and Yu, Timothy T and Ramamonjison, Rindranirina and Carenini, Giuseppe and Zhou, Zirui and Zhang, Yong},
  journal={INFOR: Information Systems and Operational Research},
  volume={62},
  number={4},
  pages={599--617},
  year={2024},
  publisher={Taylor \& Francis}
}

@inproceedings{jiang2025droc,
  title={DRoC: Elevating large language models for complex vehicle routing via decomposed retrieval of constraints},
  author={Jiang, Xia and Wu, Yaoxin and Zhang, Chenhao and Zhang, Yingqian},
  booktitle={13th international Conference on Learning Representations, ICLR 2025},
  year={2025},
  organization={OpenReview. net}
}

@misc{jiang2025rideagent,
      title={RideAgent: An LLM-Enhanced Optimization Framework for Automated Taxi Fleet Operations}, 
      author={Xinyu Jiang and Haoyu Zhang and Mengyi Sha and Zihao Jiao and Long He and Junbo Zhang and Wei Qi},
      year={2025},
      eprint={2505.06608},
      archivePrefix={arXiv},
      primaryClass={math.OC},
      url={https://arxiv.org/abs/2505.06608}, 
}

@misc{kong2026alphaopt,
      title={AlphaOPT: Formulating Optimization Programs with Self-Improving LLM Experience Library}, 
      author={Minwei Kong and Ao Qu and Xiaotong Guo and Wenbin Ouyang and Chonghe Jiang and Han Zheng and Yining Ma and Dingyi Zhuang and Yuhan Tang and Junyi Li and Shenhao Wang and Haris Koutsopoulos and Hai Wang and Cathy Wu and Jinhua Zhao},
      year={2026},
      eprint={2510.18428},
      archivePrefix={arXiv},
      primaryClass={cs.AI},
      url={https://arxiv.org/abs/2510.18428}, 
}

@misc{novikov2025alphaevolve,
      title={AlphaEvolve: A coding agent for scientific and algorithmic discovery}, 
      author={Alexander Novikov and Ngân Vũ and Marvin Eisenberger and Emilien Dupont and Po-Sen Huang and Adam Zsolt Wagner and Sergey Shirobokov and Borislav Kozlovskii and Francisco J. R. Ruiz and Abbas Mehrabian and M. Pawan Kumar and Abigail See and Swarat Chaudhuri and George Holland and Alex Davies and Sebastian Nowozin and Pushmeet Kohli and Matej Balog},
      year={2025},
      eprint={2506.13131},
      archivePrefix={arXiv},
      primaryClass={cs.AI},
      url={https://arxiv.org/abs/2506.13131}, 
}

@article{ouyang2022instructGPT,
  title={Training language models to follow instructions with human feedback},
  author={Ouyang, Long and Wu, Jeffrey and Jiang, Xu and Almeida, Diogo and Wainwright, Carroll and Mishkin, Pamela and Zhang, Chong and Agarwal, Sandhini and Slama, Katarina and Ray, Alex and others},
  journal={Advances in neural information processing systems},
  volume={35},
  pages={27730--27744},
  year={2022}
}


\appendix
\counterwithin{table}{section}
\setcounter{table}{0}

\section{LogiOR}\label{sec:logior}

An example (prob\_081) from the proposed LogiOR dataset is presented below, featuring rich annotation information. These annotations provide strong support for model evaluation and facilitate peer review to identify potential errors in the dataset, while also offering potential assistance for training reasoning models.

\begin{tcolorbox}[title=Problem description,breakable]
A large manufacturing enterprise needs to ship a consignment of 1,000 tons of goods from its factory in City F to a distribution center in City D. There are three potential transportation routes available. Each route is composed of segments using different modes of transport (road or rail). The transportation cost for each segment is calculated based on a per ton-kilometer rate. Furthermore, certain road segments are subject to a congestion fee to account for the social costs of traffic. This fee is non-linear and is calculated as \texttt{Congestion Coefficient} $\times$ (\texttt{Total Traffic Flow})$^2$. The total traffic flow on a segment is the sum of the cargo shipped by the enterprise and the existing background traffic from other companies.
The details for the three routes are provided in \cref{tab:route_details}. For congested segments, the existing background traffic is also listed.
How should the enterprise allocate the shipment of 1,000 tons of goods across the three routes to minimize the total transportation

\begin{center}
    \small
    \begin{tabular}{ccccccc}
        \toprule
        \textbf{Route} & \textbf{Segment} & \textbf{Mode} & \textbf{Distance (km)} & \textbf{Base Fee (\$/ton-km)} & \textbf{Cong. Coeff.} & \textbf{Background (tons)} \\
        \midrule
        1 & 1A & Road & 150 & 2.00 & $5\times10^{-7}$ & 2,000 \\
          & 1B & Rail & 500 & 0.80 & -- & -- \\
        2 & 2A & Road & 200 & 2.10 & -- & -- \\
          & 2B & Road & 350 & 1.90 & $8\times10^{-7}$ & 1,500 \\
        3 & 3A & Road & 100 & 2.20 & -- & -- \\
          & 3B & Rail & 600 & 0.75 & -- & -- \\
        \bottomrule
    \end{tabular}
    \captionof{table}{Route Details}\label{tab:route_details}
\end{center}
\end{tcolorbox}

\begin{tcolorbox}[title=Mathematical model,breakable]
\begin{lstlisting}[basicstyle=\ttfamily\small,breaklines=true]
Set:
1. Route 
The set of available transportation routes, \( R = \{1, 2, 3\} \)

Parameter:
1. TotalTonnage
# The total amount of goods to be shipped
1000 
2. LinearCostPerTon 
# The base transportation cost per ton for each route, excluding congestion fees. This is calculated by summing the costs of all segments for a given route (`distance * base_fee`).  
[700, 1085, 670] # in dollars per ton for Route 1, 2, and 3, respectively.
3. CongestionCoeff 
# The coefficient for calculating the congestion fee on applicable routes.  
[5e-7, 8e-7, 0] # Route 3 has no congestion fee.
4. BackgroundTraffic 
# The existing traffic volume on congested routes.  
[2000, 1500, 0] # in tons. Route 3 has no congested segment.

Decision variable:
1. TonnageOnRoute 
Continuous variable, \( TonnageOnRoute[r] \forall r \in R \), representing the amount of goods in tons transported via route \( r \).

Objective:
1. Minimize the total transportation cost. The objective is to minimize the sum of the linear transportation costs and the non-linear congestion fees for all routes.  
min: \( \sum_{r \in R} (LinearCostPerTon[r] \times TonnageOnRoute[r] + CongestionCoeff[r] \times (BackgroundTraffic[r] + TonnageOnRoute[r])^2) \)

Constraint:
1. Total Shipment Constraint. The sum of goods transported on all routes must equal the total tonnage required to be shipped.  
\( \sum_{r \in R} TonnageOnRoute[r] = TotalTonnage \)
2. Non-negativity constraint. The amount of goods transported on any route cannot be negative.  
\( TonnageOnRoute[r] \geq 0 \forall r \in R \)

Type:
Continuous, non-linear, linear  
NP
\end{lstlisting}
\end{tcolorbox}

\begin{tcolorbox}[title=Gurobipy code,breakable]
\begin{lstlisting}[basicstyle=\ttfamily\small,breaklines=true]

\begin{lstlisting}
import gurobipy as gp
from gurobipy import GRB

def solve_logistics():
    """
    Solves the transportation logistics problem with congestion pricing.
    """
    # Create a new model
    model = gp.Model("LogisticsOptimization")

    # --- Sets ---
    routes = ["Route1", "Route2", "Route3"]

    # --- Parameters ---
    total_tonnage_to_ship = 1000.0  # Total tons of goods to transport

    # Linear part of the cost for each route ($ per ton)
    linear_cost_per_ton = {
        "Route1": 700,   # (150km * $2.0) + (500km * $0.8)
        "Route2": 1085,  # (200km * $2.1) + (350km * $1.9)
        "Route3": 670    # (100km * $2.2) + (600km * $0.75)
    }

    # Congestion parameters
    congestion_coeff = {
        "Route1": 5e-7,
        "Route2": 8e-7,
        "Route3": 0  # No congestion on Route 3
    }
    background_traffic = {
        "Route1": 2000,
        "Route2": 1500,
        "Route3": 0
    }

    # --- Decision Variables ---
    # Amount of goods to ship on each route (in tons)
    tonnage_on_route = model.addVars(routes, name="TonnageOnRoute", lb=0)

    # --- Objective Function ---
    # Minimize Total Transportation Cost, Linear cost component
    total_linear_cost = gp.quicksum(
        linear_cost_per_ton[r] * tonnage_on_route[r] for r in routes
    )

    # Congestion cost component (this makes the objective quadratic)
    total_congestion_cost = gp.quicksum(
        congestion_coeff[r] *
        (background_traffic[r] + tonnage_on_route[r]) *
        (background_traffic[r] + tonnage_on_route[r])
        for r in routes if congestion_coeff[r] > 0
    )

    # Set the complete objective function
    model.setObjective(total_linear_cost + total_congestion_cost, GRB.MINIMIZE)

    # --- Constraints ---
    # 1. Total Tonnage Constraint: Must ship the exact total amount of goods
    model.addConstr(
        tonnage_on_route.sum('*') == total_tonnage_to_ship,
        name="TotalTonnageRequirement"
    )

    # Non-negativity is handled by lb=0 in variable definition.

    # Optimize the model
    model.optimize()

    # --- Print Results ---
    if model.status == GRB.OPTIMAL:
        print(f"Optimal Total Transportation Cost: ${model.objVal:,.2f}\n")
    elif model.status == GRB.INFEASIBLE:
        print("Model is infeasible. Check constraints.")
    elif model.status == GRB.UNBOUNDED:
        print("Model is unbounded. The objective can be improved infinitely.")
    else:
        print(f"Optimization ended with status {model.status}")

# Run the solver function
if __name__ == '__main__':
    solve_logistics()
\end{lstlisting}
\end{tcolorbox}

\begin{tcolorbox}[title=Problem characteristics and optimal solution,breakable]
\begin{lstlisting}[basicstyle=\ttfamily\small,breaklines=true]
"prob_081": {
    "ground_truth": 670003.8,
    "problem_type": "NP",
    "problem_size": "Toy",
    "details": {
      "variables_num": 3,
      "constraints_num": 1,
      "nonzeros_num": 3
    }
  }
\end{lstlisting}
\end{tcolorbox}

\section{Design of ORThought}
The following are the prompts for two agents of ORThought:
\begin{tcolorbox}[title=Model agent,breakable]
\begin{lstlisting}[basicstyle=\ttfamily\small,breaklines=true]
You are an expert in optimization modeling and programming. Please carefully analyze the following optimization problem:

```text
{nlp}
```

Your task is to provide a comprehensive solution that includes your detailed solution path, a formal mathematical model, and executable Gurobipy Python code. Please structure your response as follows:


Enclose your entire solution path within **<solution_path>** and **</solution_path>** tags. This section should detail your approach to understanding and modeling the problem:

1. Understanding the Problem
- **Core Optimization Objective:** What is your understanding of the primary goal of this optimization problem (e.g., what is being maximized or minimized)?
- **Key Decision Variables:**
  - Identify all the distinct choices or quantities that need to be decided.
  - For each decision variable, explain why it's a variable, its meaning in the context of the problem, and its type (e.g., continuous, integer, binary).
- **Main Constraints:** List and briefly describe the critical limitations, restrictions, or conditions imposed by the problem statement.


2. Building the Mathematical Model (Step by Step)

- **Decision Variables Definition:** Formally define each decision variable using appropriate symbols. Clearly state its meaning and mathematical type (e.g., $x_{{ij}} \ge 0$ and continuous, or $y_k \in {{0,1}}$).
- **Objective Function Construction:**
  - Clearly state whether the objective is to maximize or minimize.
  - Provide the mathematical expression for the objective function.
  - Explain the derivation of each term in the objective function, linking it directly to the problem description and the defined decision variables. Clarify how each part contributes to the overall goal.
- **Constraint Construction:**
  - For each constraint identified from the problem description:
    - Translate it into a mathematical equation or inequality involving the decision variables.
    - Explain the logic behind its formulation, ensuring it accurately reflects the corresponding limitation in the problem statement. Address aspects like fund availability, investment limits, and cash flow between years.

- **Summary of the Mathematical Model:** Compile the complete mathematical model. This section should clearly present all components of your optimization model. Enclose the entire model within **```model** and **```** tags.


3. Gurobipy Python Code

Translate your mathematical model into a complete and executable Gurobipy Python function(Everything should be defined inside of the function).
- The function has arguments **with default values extracted directly from the provided problem description**

- The function should return only the optimal objective function value if a feasible solution is found, or `None` if the problem is infeasible or unbounded.

- Enclose the Python code within **```python** and **```** tags.
\end{lstlisting}
\end{tcolorbox}

\begin{tcolorbox}[title=Solve agent,breakable]
\begin{lstlisting}[basicstyle=\ttfamily\small,breaklines=true]
You are an expert Gurobipy developer and debugger. Your task is to analyze the provided mathematical model, Gurobipy code, and error message to identify and fix the bug in the Gurobipy code. The corrected code must accurately implement the given mathematical model.

The problem description:
```text
{nlp}
```

The mathematical model:
```model
{model_text}
```

The Gurobipy code:
```code
{code_text}
```

The error message during code execution:
```text
{error_message}
```

Your Task:
1. Identify the Bug.
2. Provide Corrected Code: Offer a complete, corrected version of the Gurobipy code, and provide a brief explanation of the changes made.
3. Ensure Model Adherence: The corrected code must accurately reflect the provided mathematical model.

Output Format:
1. A brief explanation of fixes.
2. Corrected Gurobipy Code
  - Enclose the corrected code within **```code** and **```** tags.
  - The code should be a callable function whose parameters have default values and whose return value is the optimal objective function value of the model (if it exists), otherwise return None.
\end{lstlisting}
\end{tcolorbox}

\section{Setting of Ablation Study} \label{sec:appendix_c}
In this section, we introduce the settings of four ORThought variants employed in the ablation experiments, where variant specifications are implemented through prompt modifications.

\subsection{w/o repair}
Remove the component where the solve agent retrieves error information from the sandbox and conducts repairs.

\subsection{w/o understanding} 
Keep the other parts unchanged, remove the ``1. Understanding the Problem" part in ORThought.

\subsection{w/o expert knowledge in understanding} 
Keep the other parts unchanged, replace the ``1. Understanding the Problem" part in ORThought with the following content:
\begin{lstlisting}
1. Understanding the Problem
From an optimization perspective, what is your understanding of this optimization problem?
\end{lstlisting}

\subsection{w/o expert knowledge in formulation}
Keep the other parts unchanged, replace the ``2. Building the Mathematical Model (Step by Step)" part in ORThought with the following content:
\begin{lstlisting}
2. Building the Mathematical Model (Step by Step)
Please define the mathematical model. This section should clearly present all components of your optimization model. Enclose the entire model within **```model** and **```** tags.
\end{lstlisting}

\section{Examples for ORThought solution}
\subsection{A correctly solved instance}
\begin{tcolorbox}[title=Problem Description]
     A company has three industries located in different locations and needs to deliver goods to five shops in different locations. The delivery costs from the first industry to the shops are 4, 9, 2, 6, and 5, respectively, with a storage capacity of 60. The delivery costs from the second industry to the shops are 2, 6, 1, 7, and 9, respectively, with a storage capacity of 30. The delivery costs from the third industry to the shops are 2, 4, 9, 8, and 3, respectively, with a storage capacity of 45. The demands of the five shops are 15, 35, 20, 5, and 40. How can the goods be transported to minimize the cost?
\end{tcolorbox}

\begin{tcolorbox}[title=Solution Process,breakable]
\textbf{1. Understanding the Problem}

\textbf{A. Core Optimization Objective:}

The goal is to \textbf{minimize the total transportation cost} of delivering goods from three industries to five shops, respecting capacity and demand constraints.

\textbf{B. Key Decision Variables:}

\begin{itemize}
    \item \textbf{Transportation quantities:}
    For each industry \( i \in \{1,2,3\} \) and each shop \( j \in \{1,2,3,4,5\} \), define:
    \[
    x_{ij} \ge 0
    \]
    representing the amount of goods transported from industry \( i \) to shop \( j \).

    \item \textbf{Variables are continuous} because quantities can be fractional in typical transportation models.
\end{itemize}

\textbf{C. Main Constraints:}

\begin{itemize}
    \item \textbf{Supply capacity constraints:}
    The total goods shipped from each industry cannot exceed its storage capacity:
    \[
    \sum_{j=1}^5 x_{ij} \le \text{Capacity}_i
    \]
    \item \textbf{Demand fulfillment constraints:}
    Each shop's demand must be met exactly:
    \[
    \sum_{i=1}^3 x_{ij} = \text{Demand}_j
    \]
    \item \textbf{Non-negativity constraints:}
    \[
    x_{ij} \ge 0
    \]
\end{itemize}

\textbf{2. Building the Mathematical Model – Step by Step}

\textbf{Decision Variables:}
\[
x_{ij} \ge 0, \quad i=1,2,3; \quad j=1,2,3,4,5
\]

\textbf{Objective Function:}

Minimize total transportation cost:
\[
\text{Minimize} \quad Z = \sum_{i=1}^3 \sum_{j=1}^5 c_{ij} x_{ij}
\]
where \( c_{ij} \) is the cost from industry \( i \) to shop \( j \).

Given costs:
\begin{itemize}
    \item Industry 1: \( c_{1} = [4, 9, 2, 6, 5] \)
    \item Industry 2: \( c_{2} = [2, 6, 1, 7, 9] \)
    \item Industry 3: \( c_{3} = [2, 4, 9, 8, 3] \)
\end{itemize}

Total cost (expanded):
\begin{align*}
    Z = & \phantom{+} 4x_{11} + 9x_{12} + 2x_{13} + 6x_{14} + 5x_{15} \\
        & + 2x_{21} + 6x_{22} + 1x_{23} + 7x_{24} + 9x_{25} \\
        & + 2x_{31} + 4x_{32} + 9x_{33} + 8x_{34} + 3x_{35}
\end{align*}

\textbf{Constraints:}

\begin{itemize}
    \item \textbf{Supply constraints:}
    \[
    \begin{cases}
    x_{11} + x_{12} + x_{13} + x_{14} + x_{15} \le 60 \\
    x_{21} + x_{22} + x_{23} + x_{24} + x_{25} \le 30 \\
    x_{31} + x_{32} + x_{33} + x_{34} + x_{35} \le 45
    \end{cases}
    \]
    \item \textbf{Demand constraints:}
    \[
    \begin{cases}
    x_{11} + x_{21} + x_{31} = 15 \\
    x_{12} + x_{22} + x_{32} = 35 \\
    x_{13} + x_{23} + x_{33} = 20 \\
    x_{14} + x_{24} + x_{34} = 5 \\
    x_{15} + x_{25} + x_{35} = 40
    \end{cases}
    \]
    \item \textbf{Non-negativity:}
    \[
    x_{ij} \ge 0
    \]
\end{itemize}

\end{tcolorbox}

\begin{tcolorbox}[title=Mathematical Model,breakable]
\begin{align*}
    \min \quad Z = & \phantom{+} 4x_{11} + 9x_{12} + 2x_{13} + 6x_{14} + 5x_{15} \\
                  & + 2x_{21} + 6x_{22} + 1x_{23} + 7x_{24} + 9x_{25} \\
                  & + 2x_{31} + 4x_{32} + 9x_{33} + 8x_{34} + 3x_{35} \\
    \text{s.t.} \quad 
                  & x_{11} + x_{12} + x_{13} + x_{14} + x_{15} \le 60 \\
                  & x_{21} + x_{22} + x_{23} + x_{24} + x_{25} \le 30 \\
                  & x_{31} + x_{32} + x_{33} + x_{34} + x_{35} \le 45 \\
                  & x_{11} + x_{21} + x_{31} = 15 \\
                  & x_{12} + x_{22} + x_{32} = 35 \\
                  & x_{13} + x_{23} + x_{33} = 20 \\
                  & x_{14} + x_{24} + x_{34} = 5 \\
                  & x_{15} + x_{25} + x_{35} = 40 \\
                  & x_{ij} \ge 0 \quad \forall i,j
\end{align*}
\end{tcolorbox}

\begin{tcolorbox}[title=Gurobipy Code,breakable]
\begin{lstlisting}
def optimize_transportation(costs=None, capacities=None, demands=None):
    from gurobipy import Model, GRB

    # Default data based on the problem statement
    if costs is None:
        costs = [
            [4,9,2,6,5],  # Industry 1
            [2,6,1,7,9],  # Industry 2
            [2,4,9,8,3]   # Industry 3
        ]
    if capacities is None:
        capacities = [60, 30, 45]
    if demands is None:
        demands = [15, 35, 20, 5, 40]

    num_industries = len(costs)
    num_shops = len(demands)

    # Create model
    m = Model("Transportation")

    # Decision variables: x[i,j]
    # Using m.addVars for more efficient variable creation
    x = m.addVars(num_industries, num_shops, lb=0, name="x")

    # Capacity constraints
    m.addConstrs(
        (x.sum(i, '*') <= capacities[i] for i in range(num_industries)),
        name="Capacity"
    )

    # Demand constraints
    m.addConstrs(
        (x.sum('*', j) == demands[j] for j in range(num_shops)),
        name="Demand"
    )

    # Objective function
    # Using x.prod() for more efficient objective creation
    m.setObjective(x.prod(costs), GRB.MINIMIZE)

    # Optimize the model
    m.optimize()

    # Check and return the result
    if m.status == GRB.OPTIMAL:
        print(f"Optimal total cost found: {m.objVal}")
        return m.objVal
    else:
        print("No optimal solution found.")
        return None

\end{lstlisting}
\end{tcolorbox}

\subsection{Part of incorrectly solved instances}
Here, we present examples of various types of errors observed during the solution process of ORThought.

\begin{tcolorbox}[title=Example 1,breakable]
    \textbf{Problem description:}
    
    LogiCorp operates a logistics network that ships two types of products: Basic and Advanced. The raw materials needed to produce these products can be purchased for \$5 per unit. Processing 1 unit of raw material requires 2 hours of warehouse labor. Each unit of processed raw material yields 2 units of Basic Product and 3 units of Advanced Product. Basic Product can be sold for \$10/unit, and Advanced Product can be sold for \$12/unit. LogiCorp also has the option of further processing Basic and Advanced Products to produce Premium Basic and Premium Advanced, which sell for \$20/unit and \$25/unit, respectively. Each unit of Basic Product processed further requires an additional 3 hours of warehouse labor and \$5 processing cost, yielding 1 unit of Premium Basic. Each unit of Advanced Product processed further requires an additional 4 hours of warehouse labor and \$6 processing cost, yielding 1 unit of Premium Advanced. Each year, LogiCorp has 8,000 hours of warehouse labor available and can purchase up to 3,000 units of raw material. How can LogiCorp maximize its profits? Assume that the cost of warehouse labor is a fixed cost, raw materials can only be purchased in whole units.\\
    
    \noindent\textbf{Output errors:}
    
    \noindent\text{1. Understanding the Problem}
    
    \dots
    
    \noindent \text{Key Decision Variables:}
    
    \begin{itemize}

    \item Raw Material Purchase:
    
    \( R \): Number of raw material units purchased (integer, \( 0 \le R \le 3000 \))
    
    \item Processing Raw Material into Basic and Advanced Products:
    
      \( B \): Units of processed raw material producing Basic Product (continuous, \( B \ge 0 \))
      
      \( A \): Units of processed raw material producing Advanced Product (continuous, \( A \ge 0 \))
    \end{itemize}
    
    \dots
    
    \noindent\textbf{Error analysis:}
    
    \colorbox{gray!20}{\textbf{Incorrect variable}} According to the problem description, "each unit of processed raw material yields 2 units of Basic Product and 3 units of Advanced Product." This is a joint production process, meaning that processing one unit of raw material simultaneously produces both intermediate products. However, in the incorrect solution, variables $A$ and $B$ model this as a choice process, which does not align with the problem description.\\
    
    \noindent\textbf{Output errors:}
    
    \noindent\text{1. Understanding the Problem}
    
    \dots
    
    \noindent \text{Main Constraints:}
    
    \begin{itemize}
    \item Raw Material Processing:
    
      The total raw material processed into Basic and Advanced products:
        \[
        B \le 2R, \quad A \le 2R
        \]
    \end{itemize}
    
    \dots
    
    \noindent\textbf{Error analysis:}
    
    \colorbox{gray!20}{\textbf{Incorrect constraint}} The correct relationship should be "equal to", not "less than or equal to".\\
    
    \noindent\textbf{Output errors:}
    
    \noindent\text{2. Building the Mathematical Model – Step by Step}
    
    \dots
    
    \noindent \text{Objective Function:}
    
    Maximize profit, which includes revenues minus costs:

    \begin{itemize}
    \item Revenue:
        \begin{itemize}
            \item Basic Product: \( 10 \times 2B \)
            \item Advanced Product: \( 12 \times 3A \)
            \item Premium Basic: \( 20 \times P_B \)
            \item Premium Advanced: \( 25 \times P_A \)
        \end{itemize}
      \item Basic Product: \( 10 \times 2B \)
        \begin{itemize}
            \item Advanced Product: \( 12 \times 3A \)
            \item Premium Basic: \( 20 \times P_B \)
            \item Premium Advanced: \( 25 \times P_A \)
        \end{itemize}
    
    \item Costs:
        \begin{itemize}
            \item Raw materials: \( 5 \times R \)
            \item Additional processing costs: \( 5 \times P_B \) for Premium Basic, \( 6 \times P_A \) for Premium Advanced
        \end{itemize}
    \end{itemize}
    
    Profit:
    \[
    \begin{aligned}
    \text{Maximize } Z &= (10 \times 2B) + (12 \times 3A) + (20 \times P_B)\\ &\quad+ (25 \times P_A) - 5 R - 5 P_B - 6 P_A
    \end{aligned}
    \]

    \dots
    
    \noindent\textbf{Error analysis:}
    
    \colorbox{gray!20}{\textbf{Incorrect objective}} Variables $A$ and $B$ are incorrect, and therefore the objective function that calculates revenue based on them also be wrong. It fails to correctly capture the core profit logic: "the products produced from the total raw material R are partially sold directly and partially sold after further processing".\\
\end{tcolorbox}

\begin{tcolorbox}[title=Example 2,breakable]
    \textbf{Problem description:}

    A national e-commerce company operates three distribution centers in the Metro region. Each center currently stores different quantities of standard and premium products as shown in Table \ref{tab:inventory}.
    
    \noindent
    \begin{center}
    \begin{tabular}{lcc}
    \toprule
    {Center} & {Standard Products} & {Premium Products} \\ 
    \midrule
    Center 1 & 150 & 30 \\
    Center 2 & 250 & 100 \\
    Center 3 & 300 & 70 \\ 
    \bottomrule
    \end{tabular}
    \captionof{table}{Inventory levels of standard and premium products at distribution centers}\label{tab:inventory}
    \end{center}
    
    The company plans to open three retail fulfillment hubs that will serve customers with same-day delivery. Corporate policy requires that each fulfillment hub must maintain exactly 300 products in total inventory. The shipping distances (in kilometers) between distribution centers and fulfillment hubs are shown in Table \ref{tab:distances}.

    \noindent
    \begin{center}
    \begin{tabular}{lccc}
    \toprule
    From / To      & Hub 1 & Hub 2 & Hub 3 \\
    \midrule
    Center 1       & 0     & 12    & 18    \\
    Center 2       & 12    & 0     & 15    \\
    Center 3       & 18    & 15    & 0     \\
    \bottomrule
    \end{tabular}
    \captionof{table}{Center-to-hub distances (km)}\label{tab:distances}
    \end{center}
    
    What is the minimum total transportation distance required to move the products under the company's inventory requirements?\\

    \noindent\textbf{Output Errors:}

    ...
    
    Main Constraints: 
    \begin{itemize}
        \item Each hub must have the same number of premium products:
        \[
        \sum_{i} x_{ij}^P = P, \quad \forall j
        \]
        where \(P\) is the number of premium products per hub (unknown, but equal across hubs).
    \end{itemize}
    
    ...

    \noindent\textbf{Error analysis:}
    
    \colorbox{gray!20}{\textbf{Spurious constraint}} Added a spurious constraint that does not exist in the problem.
\end{tcolorbox}

\begin{tcolorbox}[title=Example 3,breakable]
    \textbf{Problem description:}

    A logistics company needs to pack 12 rectangular parcels into a shipping container with fixed width of 100 cm and variable height. The parcels have different dimensions and must be placed orthogonally (no rotations allowed). The goal is to arrange all parcels within the container while minimizing the total height used. The parcels have the following dimensions (width × height in cm): 1. 30×40, 2. 25×35, 3. 40×20, 4. 15×25, 5. 35×30, 6. 20×15, 7. 25×20, 8. 30×25, 9. 15×10, 10. 20×30, 11. 10×15, 12. 25×10. Parcels cannot overlap and must be completely contained within the container's width. The container has unlimited height capacity, but the objective is to minimize the used height.\\

    \noindent\textbf{Output Errors:}
    
    \noindent\text{1. Understanding the Problem}
    
    \dots
    
    \noindent \text{Key Decision Variables:}
    \begin{itemize}
        \item Overlap: Parcels cannot overlap.
    \end{itemize}
    
    \dots
    
    \noindent \text{Main Constraints:}
    \begin{itemize}
        \item Non-overlapping:
      For any two parcels \( i \neq j \), either parcel \( i \) is to the left of parcel \( j \), or to the right, or above or below, which can be modeled with binary variables.
    \end{itemize}
    
    \dots

    \noindent\text{2. Building the Mathematical Model – Step by Step}\\
    \text{Decision Variables:}\\
    \begin{itemize}
        \item \( z_{ij} \in \{0,1\} \): binary variables indicating relative positioning between parcels \( i \) and \( j \).
    \end{itemize}
    
    \dots

    \noindent \text{Constraints:}
    \begin{itemize}
        \item Non-overlap constraints:\\
        For each pair \( (i, j) \), \( i < j \):
        \[
        x_i + w_i \leq x_j + M (1 - z_{ij})
        \]
        \[
        x_j + w_j \leq x_i + M z_{ij}
        \]
        where \( M \) is a large constant (e.g., 1000) to deactivate constraints depending on \( z_{ij} \).
        
        Additionally, enforce that for each pair, either \( i \) is to the left of \( j \) or vice versa:
        \[
        z_{ij} + z_{ji} = 1
        \]
        with \( z_{ji} = 1 - z_{ij} \).
    \end{itemize}
    
    \dots

    \noindent\textbf{Error analysis:}
    
    \colorbox{gray!20}{\textbf{Missing variables \& Missing constraints}} During the Understanding the problem phase, the solve agent identified that non-overlapping constraints should be added for parcels in two directions. However, during the Building the mathematical model stage, this was not strictly implemented — only a single-direction relationship variable $ z $ was introduced, and only non-overlapping constraints in one direction were added. The variables and constraints that represent parcel relationships in the other direction were missing.
\end{tcolorbox}
\end{document}